**Exploring the potential of AlphaEarth and TESSERA embeddings for Fine-scale Local Climate Zone Mapping: A case study across five cities in Switzerland**

Htet Yamin Ko Ko[1,*], Clement Atzberger[2]

1. International Space Science Institute, Bern, Switzerland
2. CYCLOPS MRV, United States

*Corresponding authors: koko.htetyamin@issibern.ch

Abstract

Understanding urban spatial morphology is critical for climate modeling, risk assessment, and sustainable urban design, and Local Climate Zone (LCZ) mapping provides the basic framework for this. However, many cities still use coarse ~100-m resolution LCZ records, which are unsuitable for fine-scale urban research. In this study, precomputed embeddings from TESSERA (Feng et al., 2025) and AlphaEarth (Brown et al., 2025) are compared to traditional Sentinel-1/2 (S1S2) composites in five Swiss cities to see if they can upscale coarse LCZ maps to 10-m resolution using an attention-based U-Net. Three experiments assess multi-city transferability, the impact of higher-resolution reference data, and temporal robustness to year-to-year phenology changes. We find that all datasets achieve strong performance with test data Intersection-over-Union (IoU) ranging from 0.59-0.69 and 0.77-0.82 in the first two experiments. TESSERA consistently outperforms both S1S2 and AlphaEarth across both settings As expected, we find that the transfer of embedding-based models from one year to another remains an open challenge. Overall, however, our results demonstrate the promising potential of embeddings derived from EO foundation models to reduce time consuming preprocessing, respectively, manual feature engineering tasks and to guide a universal deep learning-based LCZ mapping workflow. When combined with a simple location-aware attention U-Net architecture, the embeddings enhance regional transferability and scalability, supporting the development of comprehensive and reproducible fine-scale LCZ maps for global urban climate applications Improving reference data quality remains the strongest lever for further accuracy gains.

## 1. Introduction

Cities are exposed to the emerging urban heat and accelerating impacts of global climate change (Demuzere et al., 2022). Consequently, cities are experiencing hotter temperatures than rural regions and there is an increasing risk of hospitalization and fatality amid urban vulnerable populations (Wellinger et al., 2024). These effects are most likely to further increase in the future, as the urban population is growing at a very fast pace, while the climate continues to change (Wellinger et al., 2024). In the mean times, cities are also a major driver of regional and local climate changes due to their emissions of air pollutants, anthropogenic heat release, high energy consumption, and the prevalence of impervious surfaces with subsequent changes of the energy and water balance (Demuzere et al., 2022).

Hence, understanding urban spatial morphology including build-up structure and material composition has become essential for urban climate modelling, risk assessment and sustainable urban planning (Wellinger et al., 2024). Classical ‘Land Use Land Cover’(LULC) maps can provide valuable information about usage and coverage of land but often lack the detailed urban-specific information needed to better model the microclimate within urban landscapes. To address this limitation, 'Local Climate Zone' (LCZ) classification schemes are increasingly seen as a suitable standardized framework to characterize urban ecosystems according to physical (3D) structure, coverage, human activity and material composition (Wellinger et al., 2024).

As LCZ characterizes urban morphology in a more appropriate way compared to simple LULC maps, LCZs have become a fundamental feature of urban climate research, notably in studies exploring the urban heat island effect, but also with respect to the general urban thermal situation (Moix & Giuliani, 2024). Considering the benefits of LCZ mapping, it is expected to see numerous research developing LCZ maps to support sustainable urban development [2]. Since LCZ maps can greatly support evidence-based environmental studies, the need of fine-scale spatial information with respect to urban landscapes is growing (Moix & Giuliani, 2024).

While the value of LCZ maps is well recognized, to date, many cities still rely on the WUDAPT 100-m spatial resolution globally available LCZ dataset established by Demuzere et al (Demuzere et al., 2022). While overall very useful, given the small-scale structure of most cities, the 100-m scale is frequently too broad to provide the necessary spatial detail. With this stated gap, the aim of this study is to investigate whether coarse resolution LCZ data together with open-source, ready-to-use (precomputed) embedding datasets can be leveraged to generate fine-scale (10 m resolution) LCZ maps. The proposed study is facilitated on one hand by the availability of precomputed embeddings at deca-metric resolution, as well as recent advances in deep learning, and the availability of more powerful computational resources. Together, these recent developments and datasets provide new opportunities to tackle the necessary fine-scale mapping of local climate zones.

Few studies have applied deep learning approaches adapted specifically to LCZ mapping frameworks (Zhu et al., 2024).

In this study, we inspect open-source satellite datasets with 10 m spatial resolution and benchmark their capabilities to generate LCZ mapping using the coarse resolution WUDAPT 100 m global LCZ dataset for model training. We employ an attention-based U-Net architecture designed for multi-channel input images and evaluate how well the geospatial information of seasonal composites (S1S2), respectively, embedded datasets can be translated into high-resolution LCZ maps. With respect to the embedding datasets, we have selected two of the leading global datasets provided by TESSERA (Feng et al., 2025) and AlphaEarth (Brown et al., 2025).

The study aims to present and evaluate a unified U-Net-based deep learning approach for producing fine-scale LCZ maps at 10m spatial resolution for five Swiss cities, based on reference data extracted from the 100-meter WUDAPT LCZ map (Demuzere et al., 2022) and utilizing as predictive feature precomputed embeddings (Feng et al., 2025; Brown et al., 2025). By utilizing training labels available at global scale, the proposed model can potentially be applied to generate LCZ maps for any city in any country. To assess if embeddings provide advantages compared to a more classic data processing pipeline, a separate baseline model was developed. The baseline model uses S1S2 composites instead of precomputed embeddings with otherwise identical processing, U-Net architecture and analysis (***Figure 1***). To assess if higher resolution reference labels provide an improved classification accuracy, we test in a second experiment the use of reference labels at 78 m. This analysis is restricted to the city of Bern for which such higher resolution reference information is available. In a third experiment, we test the temporal transferability of models from one year to another. By making the developed python workflow open source, we allow other users to generate revised LCZ maps of their own region, potentially using locally available reference labels of higher quality and/or with more thematic depth.

## 2. Objectives

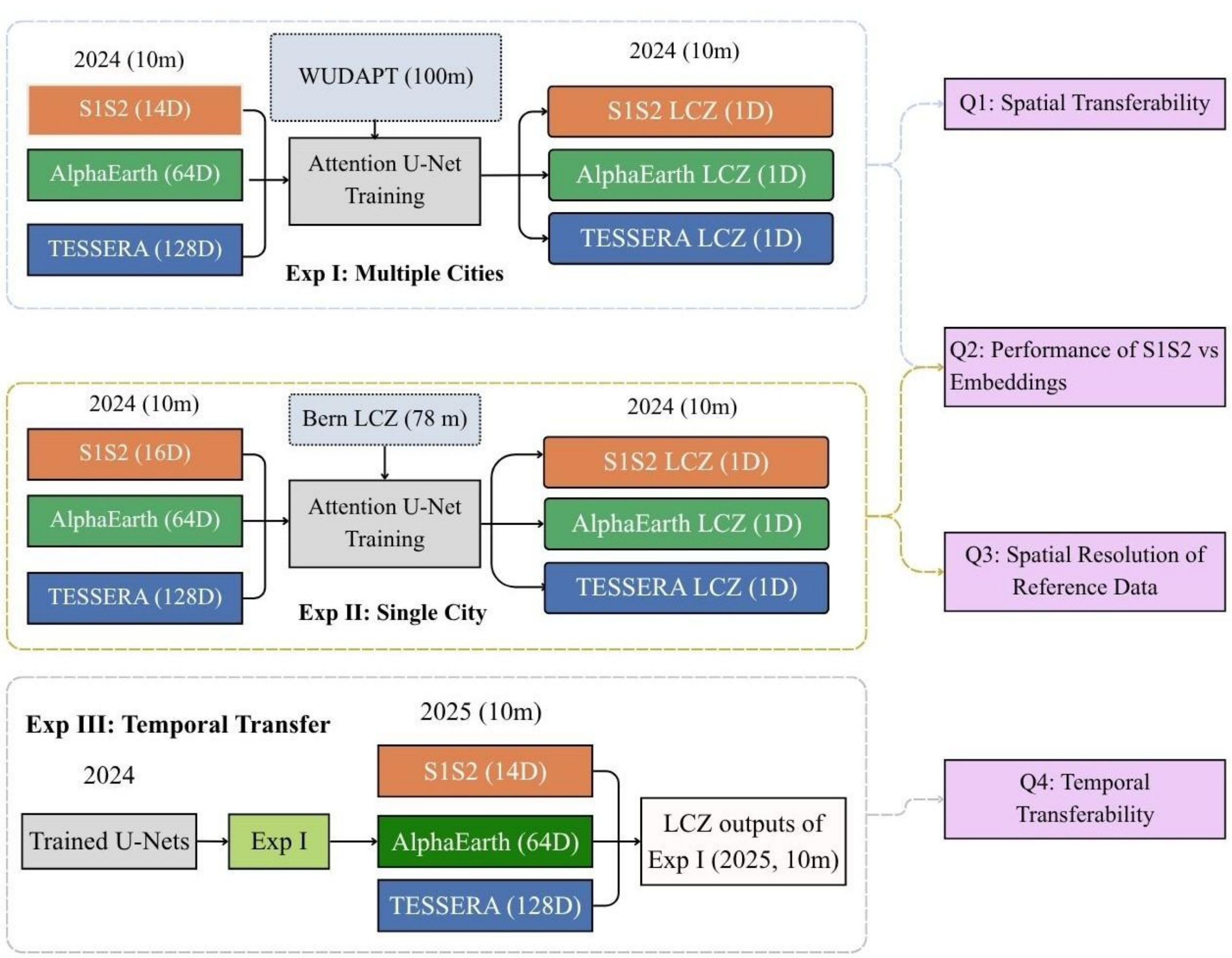

**Figure 1**: General overview of the three experiments and showing the research questions tackled in this study. For each input dataset a different Attention U-Net model is trained to handle differences in the input dimension. The first experiment (I) derives LCZ maps for five cities using a single model trained on globally available reference labels at 100m. In the second experiment (II) the potential benefits of higher resolution reference data are assessed. This experiment is restricted to a single city (Bern) for which local reference data at 78m resolution is available. In the third experiment (III), the U-Net trained in Exp I on 2024 data is applied to data of the following year (2025). This allows us to assess the robustness of models to year-to-year changes in land surface phenology.

The success of our approach is assessed using four user-centric criteria:

- Performance: A good LCZ map must agree with the ground situation expressed through high accuracy and 'Intersection over Union' (IoU) scores. As the accuracy of the WUADPT global LCZ map used for training purposes can vary based on the regions, we acknowledge that the performance of our developed framework may be conditioned on the quality of the reference labels.
- Spatial Transferability: As globally available datasets are used as inputs/targets for this developed framework, the framework is inherently designed for cross-regional application. Its spatial transferability is evaluated across five major Swiss cities, demonstrating its capability to expand beyond individual urban areas (while being still restricted to a single country).
- Accessibility: The embeddings-based workflow decreases the need for manual ground sample data collection, data preprocessing and feature engineering. By incorporating open-source (reference and predictor) datasets and a simple U-Net-based deep learning architecture, the framework reduces computational and technical barriers. As a result, the development of fine-scale Local Climate Zone (LCZ) maps is becoming increasingly feasible for the global research community as well as other stakeholders.
- Temporal Transferability: We also evaluate if the trained model can be successfully used to generate LCZ maps in years not trained on, thereby assessing the models/embeddings temporal transferability.
- Updating frequency: By utilizing readily available embeddings as the main input for LCZ mapping, we can track more easily how LCZ/urban changes over time rather than relying on infrequently updated LCZ maps. Both, the producers of TESSERA and AlphaEarth embeddings have committed to annually update the global embedding datasets.

## 3. Study Area and Datasets

Recent developments in satellite-based 'geospatial foundation models' (FMs) provide a promising approach for consistent, large scale and low-cost environmental monitoring purposes (Brown et al., 2025; Chaiyana et al., 2026; Feng et al., 2025; Hamoudzadeh et al., 2026; Houriez et al., 2025). As satellite embeddings tend to become the new standard of analysis-ready geospatial products, it is essential to evaluate their suitability for urban studies. In this study, we evaluate two of the leading embedding-based datasets for their suitability for Local Climate Zone (LCZ) mapping and compare their performance with a baseline model using traditional Sentinel datasets and multi-spectral composites.

### 3.1. Study Area

Switzerland is selected as the research location, and the five major Swiss cities have been selected for the purpose of this study (**Figure *2***): Bern, Basel, Geneva, Lausanne and Zurich. A short description of each city is given below, and important information is summarized in **Table 1**.

Amongst the five selected cities, Bern is the smallest city with a population of nearly 135,000, which results in a population density of 2600 inhabitants per km$^2$ (Wellinger et al., 2024). For the study area of Bern, a rectangle was defined spanning 13.4 km by 11.7 km covering Bern and its neighbouring municipalities (Wellinger et al., 2024). Bern was selected as it allows to study the impact of reference data quality. Indeed, Bern is the only city for which reference data at 78m is available compared to the 100m for the remaining cities.

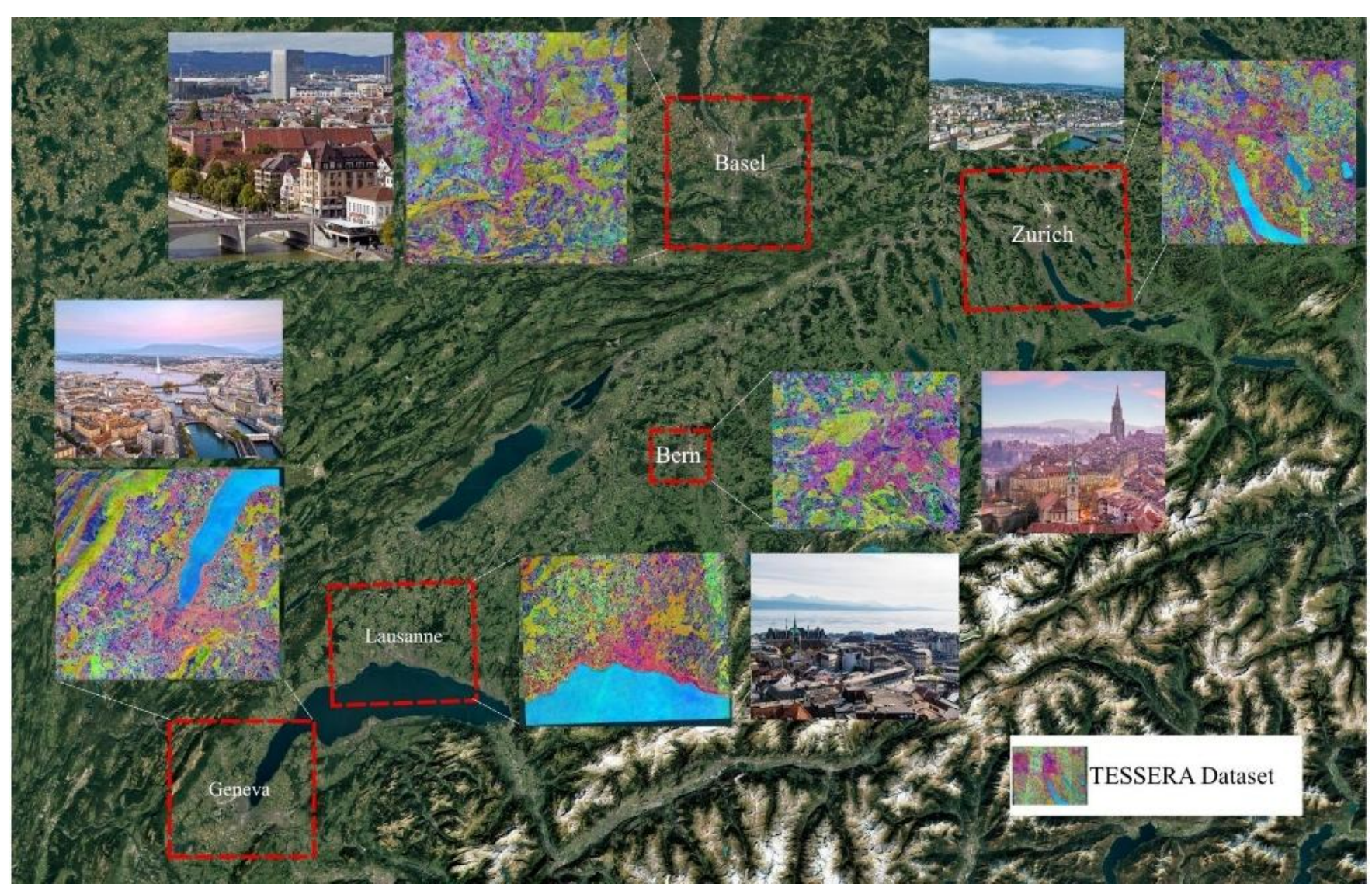

**Figure 2** Study area with the five biggest cities of Switzerland: Basel, Bern, Geneva, Lausanne, and Zurich. The background true colour image is from (Google, 2024). The displayed embeddings are from TESSERA.

Zurich is the top most populated city in Switzerland, with more than 400,000 residents. It is in the northeast on the Swiss plateau by Lake Zurich. Zurich's urban landscape, which is encircled by hills up to 900 meters high, is made up of a combination of contemporary residential buildings on the periphery and dense historic buildings in the city center (Moix & Giuliani, 2024). A ROI rectangle was defined to capture the city and its surrounding area with size 28.6 km by 24.0 km covering the whole city and its neighboring communities.

Geneva, Switzerland's second largest city, has over 200,000 inhabitants and is in the country's southwest along Lake Geneva. It has a dense historic core and modernized residential areas in the neighboring communities (Moix & Giuliani, 2024). For Geneva, a ROI rectangle is defined as 30.8 km by 28.3 km.

With over 173,000 residents, Basel is the third most populous city in Switzerland. It is in the north and borders both France and Germany. The historic core of Basel gives way to modern residential constructions on the outskirts (Moix & Giuliani, 2024). Basel’s study area was defined by spanning 20.4 km by 26.0 km covering the entire city and its neighbouring municipalities (Moix & Giuliani, 2024).

Situated on the northern beaches of Lake Geneva, Lausanne is the fourth largest city in Switzerland, home to over 140,000 people. Built on hills, the city has a historic center encircled by modern residential areas (Moix & Giuliani, 2024). For the study area of Lausanne, a rectangle was defined spanning 30.7 km by 23.6 km covering the entire city of Lausanne and its neighbouring municipalities (Moix & Giuliani, 2024).

***Table 1**: Characteristics of the five selected cities in this study in alphabetical order*

| | Basel | Bern | Geneva | Lausanne | Zurich |
|---|---|---|---|---|---|
| Lat/Lon of City Center (EPSG:4326) | 47.55 N, 7.59E | 46.94 N, 7.44E | 46.2N, 6.15E | 46.51N, 6.63E | 47.37N, 8.54E |
| ROI area ($km^2$) | 691.0 | 156.7 | 871.8 | 718.6 | 687.9 |
| Description (Moix & Giuliani, 2024) | Third largest city located in the northwestern part along the Rhine River near the borders of France and Germany | Federal City of Switzerland and fifth largest city with smallest study area in this study | Second largest city situated along Lake Geneva in the southwest of the country | The fourth-biggest city on Lake Geneva's northern shores | Largest Swiss city located near Lake Zurich on the Swiss plateau in the northeast |
| Urban Structure(Moix & Giuliani, 2024) | Historic city center along with dense residential and commercial zones | Historic city center along with dense residential areas in the old town region | Modern residential areas surround a densely populated historic center | Located on hills, it has a historic core encircled by modern residential structures. | A combination of modern residential buildings on the fringes and a dense concentration of old buildings in the center |
| Topography (Moix & Giuliani, 2024) | Surrounded by Rhine River | Surrounded by hills of up to 858 m, and crossed by Aare River | Crossed by Rhone and Arve rivers | Located on hilly landscape | Surrounded by hills of up to 900 m |
| Climate (Köppen) | Cfb Temperate Oceanic Climate | | | | |
| Annual Mean temperature (°C)/ precipitation of 2025 (mm) (Meteoswiss, 2026) | 11.8 /841.9 | 10.3/1021.8 | 12.0/945.7 | 10.8/853.7 | 10.6/1107.9 |
| Population | 173,000 | 135,000 | 200,000 | 140,000 | 400,000 |
| Population density ($km^{-2}$) (Federal Statistical Office, 2024; Wiedmann et al., 2023) | 7 445 | 2 673 | 13 099 | 3 502 | 4 965 |
| GDP per Capita (CHF Billion) (Federal Statistical Office, 2022) | 41 | 89 | 61 | 65 | 164 |
| S1 Relative Orbit Number | [88,139,15,66] | [88,139,15,66] | [88,139,161,37] | [88,139,161] | [88,139,15,66] |
| S2 tile(s) | 32TMT, 32TLT | 32TLS, 32TLT | 31TGM | 31TGM, 32TLS | 32TMT |
| Google AlphaEarth Tile(s) | xcnalm3la6maroyge | xlhcaokskggk40mla | x7wbue81rzj8u1di6, xjmzcenb18fy6g8he | x7wbue81rzj8u1di6 | xcnalm3la6maroyge, xlhcaokskggk40mla |

### 3.2. Reference LCZ Dataset

Two reference LCZ datasets were used in this study: WUDAPT (Demuzere et al., 2022) and Bern LCZ (Wellinger et al., 2024). The latter reference dataset has a better spatial resolution (78 m instead of 100 m) but is only available for the city of Bern. It was included here to assess potential implication of the reference data resolution on the derived LCZ quality.

#### 3.2.1. WUDAPT LCZ Dataset: Basel, Bern, Geneva, Lausanne and Zurich

WUDAPT stands for "World Urban Database and Access Portal Tools" and develops LCZ maps based on crowdsourced information (Wellinger et al., 2024). This dataset is produced at 100 m and was used for all cities in Exp I and III. The WUDAPT maps classify the urban and peri-urban landscape into 17 standardized classes and refer to the year 2019 (Demuzere et al., 2022). Note that several classes are not present in the five investigated cities, so that the final thematic depth in Exp I and III was 13 classes (**Table 2**a).

#### 3.2.2. Bern LCZ Map

***Table 2a:*** *Original WUDAPT LCZ code and presence of the various classes in the five Swiss cities for Exp I and III. Note that four of the 17 classes are not present in any of the studied cities leading to a thematic depth of 13 classes. Note that the reference WUDAPT LCZ dataset*

*is 100 m resolution.*

| **WUDAPT LCZ** | **Class Name** | **Classes present for Exp I and III** | | | | |
|---|---|---|---|---|---|---|
| | | Basel | Bern | Geneva | Lausanne | Zurich |
| 1 | Compact high-rise | | | - | | |
| 2 | Compact mid-rise | • | • | - | - | • |
| 3 | Compact low-rise | • | - | • | • | • |
| 4 | Open high-rise | - | | | | |
| 5 | Open mid-rise | • | • | • | • | • |
| 6 | Open low-rise | • | • | • | • | • |
| 7 | Lightweight low-rise | - | | | | |
| 8 | Large low-rise | • | • | • | • | • |
| 9 | Sparsely built | • | • | • | • | • |
| 10 | Heavy industry | - | | | | |
| A | Dense trees | • | • | • | • | • |
| B | Scattered trees | • | • | • | • | • |
| C | Bush / scrub | • | | • | • | • |
| D | Low plants | • | • | • | • | • |
| E | Bare rock / paved | • | • | • | - | • |
| F | Bare soil / sand | | | • | | - |
| G | Water surface | • | • | • | • | • |

Higher resolution (78 m) ground truth labels for model training and evaluation of Bern were derived from data provided by the Urban Climate Group of University of Bern . This dataset is further indicated as "Bern LCZ" (2023). The Bern LCZ map is based on WUDAPT data but further refined by using measurable physical properties and local knowledge including urban canopy layers, building height, sky view factor, surface cover fraction, and anthropogenic heat flux (Wellinger et al., 2024). Despite the refinement, only 14 classes are distinguished as three LCZ classes are not present in Bern: LCZ 3 (*Compact high-rise*), LCZ 7 (*Lightweight low-rise*) and LCZ F (*Bare soil / sand*) (Wellinger et al., 2024) (**Table 2**b).

**Table 2b:** Original WUDAPT LCZ code and presence of the various classes in Bern for Exp II. Note that three of the 17 classes are not present in Bern, leading to a thematic depth of 14 classes. Note that the reference Bern LCZ dataset is 78 m resolution.

| WUDAPT LCZ | Class Name | Classes present for Exp II |
|---|---|---|
| | | Bern |
| 1 | Compact high-rise | ● |
| 2 | Compact mid-rise | ● |
| 3 | Compact low-rise | - |
| 4 | Open high-rise | ● |
| 5 | Open mid-rise | ● |
| 6 | Open low-rise | ● |
| 7 | Lightweight low-rise | - |
| 8 | Large low-rise | ● |
| 9 | Sparsely built | ● |
| 10 | Heavy industry | ● |
| A | Dense trees | ● |
| B | Scattered trees | ● |
| C | Bush / scrub | ● |
| D | Low plants | ● |
| E | Bare rock / paved | ● |
| F | Bare soil / sand | - |
| G | Water surface | ● |

### 3.3. Alpha Earth Foundation Dataset

AlphaEarth foundation data is developed by Google (Brown et al., 2025). It provides global, annual 10-meter resolution satellite-derived embeddings for the years 2017 to 2025 (Brown et al., 2025). The freely available dataset merges multisource observation of available global geospatial datasets (optical, radar, lidar and climate datasets) and embeds this information into 64 layers at 10-meter resolution. The approach for learning the embeddings is semi-supervised with a strong focus on masked autoencoders for learning. Data are processed in mini-patches aiming to incorporate spatial structure into the embeddings. The direct accessibility of AlphaEarth embeddings through the 'Google Earth Engine' (GEE) platform, and the availability of annual dataset from 2017 to 2025 with global coverage (Brown et al., 2025), are major drivers for the increased usage of this dataset.

## 3.4. TESSERA dataset

TESSERA is the acronym of 'Temporal Embeddings of Surface Spectra for Earth Representation and Analysis' (Feng et al., 2025). The dataset at 10m resolution is currently available for the years 2024 (global) and 2025 (Europe) as well as for requested regions, with ongoing processing aiming to ensure global coverage at annual resolution from 2017-2025. TESSERA provides yearly gap-free embeddings with 128 bands at 8-bit quantization and 10-meter resolution. TESSERA embeds time series of Sentinel-1 (microwaves) and Sentinel-2 (optical) through a fully self-supervised learning approach based on the Barlow Twins loss, forcing the outputs to become agnostic against missing observations while ensuring uncorrelated features (Lisaius et al., 2024; Zbontar et al., 2021). TESSERA provides analysis-ready outputs by using only spectral-temporal features at pixel level (Feng et al., 2025), contrasting with AlphaEarth' patch-based approach (Brown et al., 2025).

## 3.5. S1S2 EO time series and seasonal composites

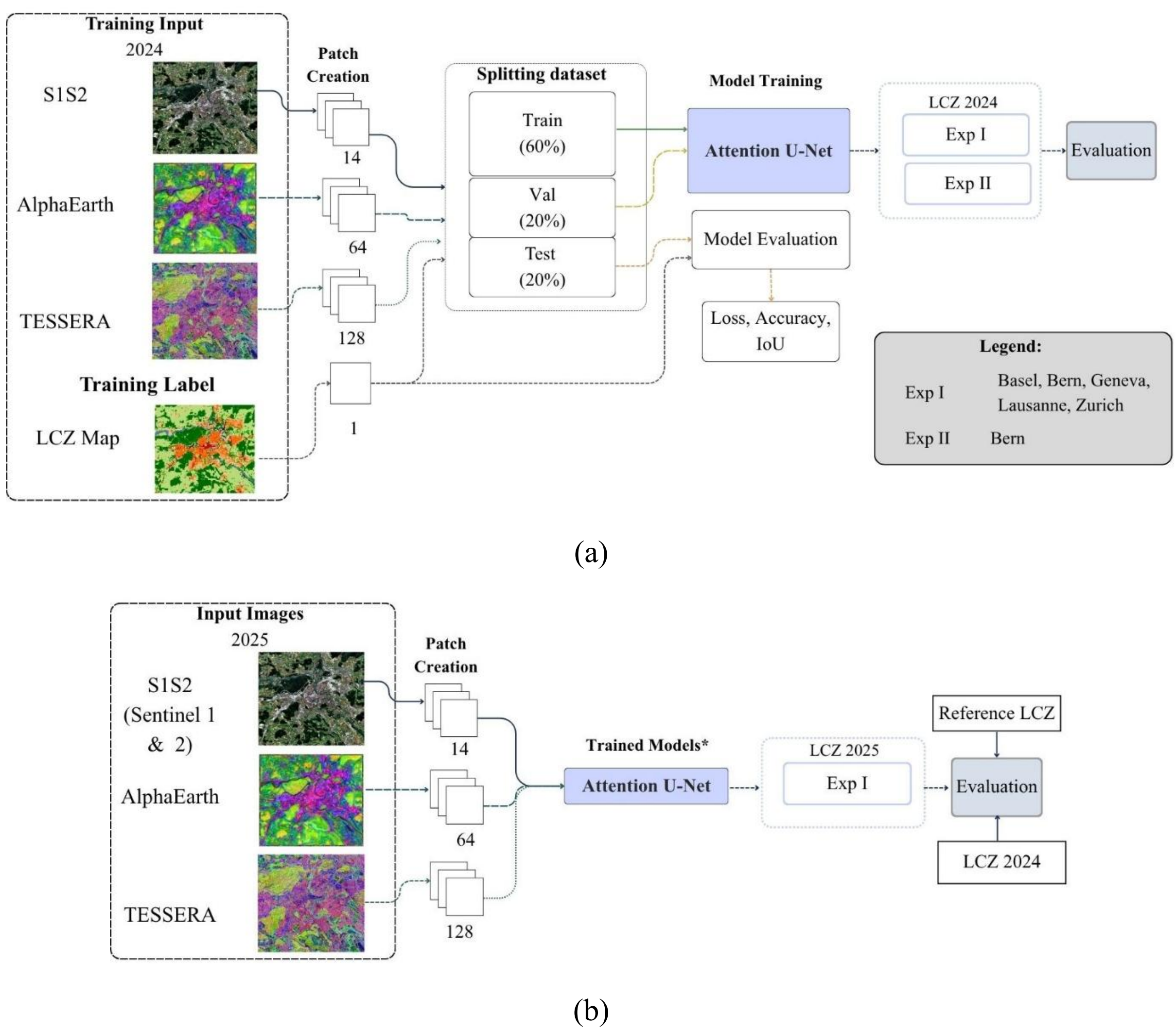

*Figure 3: Illustration showing the different experiments presented in this study and the datasets with varying dimensionality involved. The upper part (a) shows the general learning approach which allows us to infer LCZ based on remotely sensed inputs and coarser resolution reference maps. Two experiments are conducted: (i) a U-Net is trained and applied*

*across multiple cities using reference data at 100m, and (ii) a single city is analyzed but using reference data at 78m. Both experiments involve the comparison of three input training datasets: seasonal composites ("S1S2") and two precomputed embeddings (AlphaEarth and TESSERA). The lower part (b) shows the analysis conducted with respect to the temporal transfer. Here, U-Nets are trained on one year of data (2024) and applied/assessed at the following year (2025). * indicates that the U-Nets from the first two experiments have been frozen and applied to the new predictor sets.*

The freely available Landsat and Sentinel satellite missions remain critical and fundamental not only as inputs of most geospatial foundation models but also for more traditional LULC mapping and environmental analysis. Caused by their global coverage, high temporal revisit frequency, and long-term historical record, these datasets are used at steadily increasing pace (Chaiyana et al., 2026). However, working with such data requires extensive preprocessing such as cloud removal, atmospheric correction, design and selection of indices and features. None of these tasks are trivial but fundamental to enable sufficient accuracy of the subsequent analysis (Chaiyana et al., 2026).

Thanks to the emerging technology of satellite-derived embeddings, users can in principle skip the extensive preprocessing requirements of traditional satellite datasets and directly work with the embeddings. However, the research community is not yet at the stage to take an informed decision. To justify the use of embedding instead of the original time series, it must be proven that the accuracy of the downstream task can at least be maintained. S1S2 imagery serves therefore in this study as a baseline, against which the embeddings-based approaches can be compared. The workflow designed for this comparison is shown in **Figure 3** together with the use of embeddings.

For Sentinel 1, S1 Ground Range Detected (GRD) scenes are used. The data are processed using the Sentinel-1 Toolbox to generate a calibrated, ortho-corrected product (Google Earth Engine, n.d.; Houriez et al., 2025). Data from both ascending and descending orbits with dual-band cross-polarization, vertical transmit/horizontal received (VV and VH) (total four bands), are selected. Four median composites (June-August) are created per year and used in the analysis.

For Sentinel 2, Level-2A orthorectified atmospherically corrected surface reflectance dataset from GEE data catalog is used (Chaiyana et al., 2026; Google Earth Engine, n.d.). Ten spectral bands from the Sentinel 2 MSI dataset are selected, incorporating bands from visible, red edge, near infrared and short-wave infrared (Chaiyana et al., 2026). Again, a seasonal median (2024) composite (June-August) is created per spectral band for further analysis. This yields additional 10 features for the baseline model. **Table 3** provides an overview of all datasets.

**Table 3**: Description of the different datasets used in the various experiments: S1S2, AlphaEarth, and TESSERA. Experiment I is the multi-city study (2024); Experiment II is the single city study using higher resolution reference data (2024); Experiment III is the multi-city temporal transferability study where trained models using data from 2024 are applied to predict datasets of 2025. Note that the different input datasets involve different number of spectral-temporal features: S1S2: 14-16 features, AlphaEarth: 64 features and TESSERA: 128 features. The number of features for the S1S2 data varies between experiments as the two topographic

features (DSM and DEM) are not available for all cities and hence have only been used in Experiment II.

| | Dataset | Features | Collection Time | Platform | Experi-ment |
|---|---|---|---|---|---|
| S1S2 | Sentinel-1 Dual-polarization C-band Synthetic Aperture Radar (SAR) | 4 features VV and VH from ascending and descending orbits, respectively | 3-month median composite (June-August 2024) | Google Earth Engine | I, II, III |
| | Sentinel-2, Level-2A surface reflectance | 10 Features (B2, B3, B4, B5, B6, B7, B8, B8A, B11, B12) | | | I, II, III |
| | Swiss Surface3D (Federal Office of Topography swisstopo, 2025) | Digital Surface Model | 2023 | Swiss topography | II |
| | Swiss TLM3DRegio | Digital Elevation Model | 2025 | | II |
| AlphaEarth | Embeddings | 64D embeddings derived from annual EO time series | 2024 | Google Earth Engine | I, II, III |
| TESSERA | Embeddings | 128D embeddings derived from annual EO time series | 2024 | Geotessera Python Package | I, II, III |

### 3.5.1. Digital Terrain Model (DTM) and Digital Surface Model (DSM) from Federal Office of Swiss Topography

The Swiss Surface 3D model (DSM) (Federal Office of Topography swisstopo, 2024) and topographic landscape model (DTM) (Federal Office of Topography swisstopo, 2025) datasets are used as fifteen and sixteen bands of the baseline dataset for Experiment II (**Table 3**). These datasets are derived from airborne LiDAR data in 2023 (DSM) and 2025 (DTM), respectively. Since Basel and Geneva are not entirely located on Swiss territory, DSM and DTM are not accessible for full coverage. Hence, for Experiment I, only fourteen features (Sentinel 1 and 2) are used as baseline dataset (**Table 3**).

## 4. Methodology

Three experiments were conducted to derive LCZ maps from remotely sensed inputs (**Figure 3**): (i) analyzing all five major Swiss cities using a single U-Net (Exp I), (ii) using only Bern with higher resolution reference label for analysis (Exp II), and (iii) checking the temporal transferability (Exp III).

The purpose of adding multiple cities in Experiments I and III is to evaluate the framework's ability to scale easily and accurately across different urban morphologies, respectively, to assess temporal model transferability. The two experiments have a thematic depth of 13 classes (**Table 2a**). Experiment II was performed to evaluate the effect of reference label resolution on classification accuracy. This resulted in a thematic depth with a total of 14 LCZ classes (***Table 2***b).

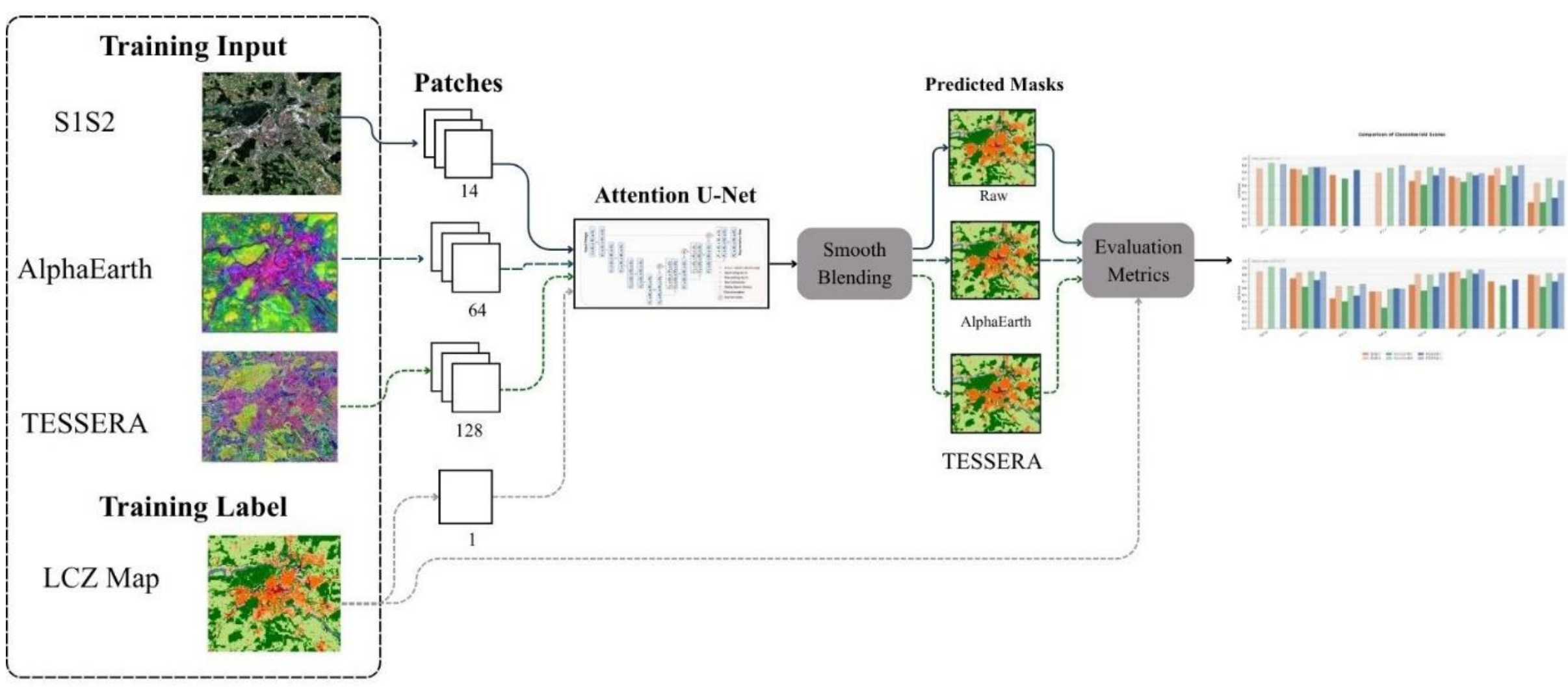

**Figure 4**: Overview of the proposed approach for model training/validation and testing. The illustration presents the workflow of Exp I, and the size of patches are 128 × 128 × number of features from each dataset. Three U-Net models are created for three different datasets in Experiment I. The outcome of Exp I is the LCZ maps of five cities.

The LCZ mapping methodology (**Figure 4**) can be split into five steps (details are given in sections 5.1 to 5.3):

1) **Data preprocessing.** The first step is the training data preparation, where we prepared the reference LCZ maps of different Swiss cities and patched the images to 128 x 128 size for the deep learning algorithm. Raster alignment between input images and the reference data is performed to ensure perfect alignment at the same resolution.
2) **Generation of training data.** In the second step, the created input and label patches are analyzed to remove patches with a single dominant LCZ class. This step is incorporated to ensure a balanced representation of LCZ types in the training process.
3) **Attention U-Net training.** In the third step, the pre-processed dataset is used to train a modified attention U-Net Architecture. The architecture was chosen after internal benchmarking multiple U-Net Architecture backbones such as ResNet50, ResNet18, and U-Net with ImageNet weights.
4) **Inference and map generation.** In the fourth step, the different feature sets

(composites and embeddings, respectively) are patched into the same size as used for the training and passed into the best performing models. The output patches are tiled back to original shape by using the smooth blending of patches to minimize edge effects (Bhattiprolu, 2020; Guillaume Chevalier, 2017).

5) **Statistical evaluation.** To evaluate the resulting LCZ maps and the usability of the framework, two types of evaluation are conducted: (i) model-specific accuracy assessment, and (ii) class-wise area computation and comparison with reference label datasets. The statistical evaluation was done on a test dataset separated from the training and validation data (**Figure 3a**).

## 4.1. Preprocessing

### 4.1.1. Spatial Alignment to Embedding Grid

To enable pixel-wise alignment between the different resolutions of LCZ reference labels and the input embedding satellite images, the 78 m (Bern LCZ) and 100 m (WUDAPT LCZ) reference LCZ raster were reprojected and resampled to match the spatial grid of the satellite data at 10 m spatial resolution. Reprojection was performed with nearest-neighbor resampling, and the reference coordinate system is the LV95 coordinate system (Federal Office of Swiss Topography, 2024). The aligned label raster ensures the same affine transform, pixel grid origin, and spatial extent as the input embedding tiles.

### 4.1.2. Training Dataset Creation

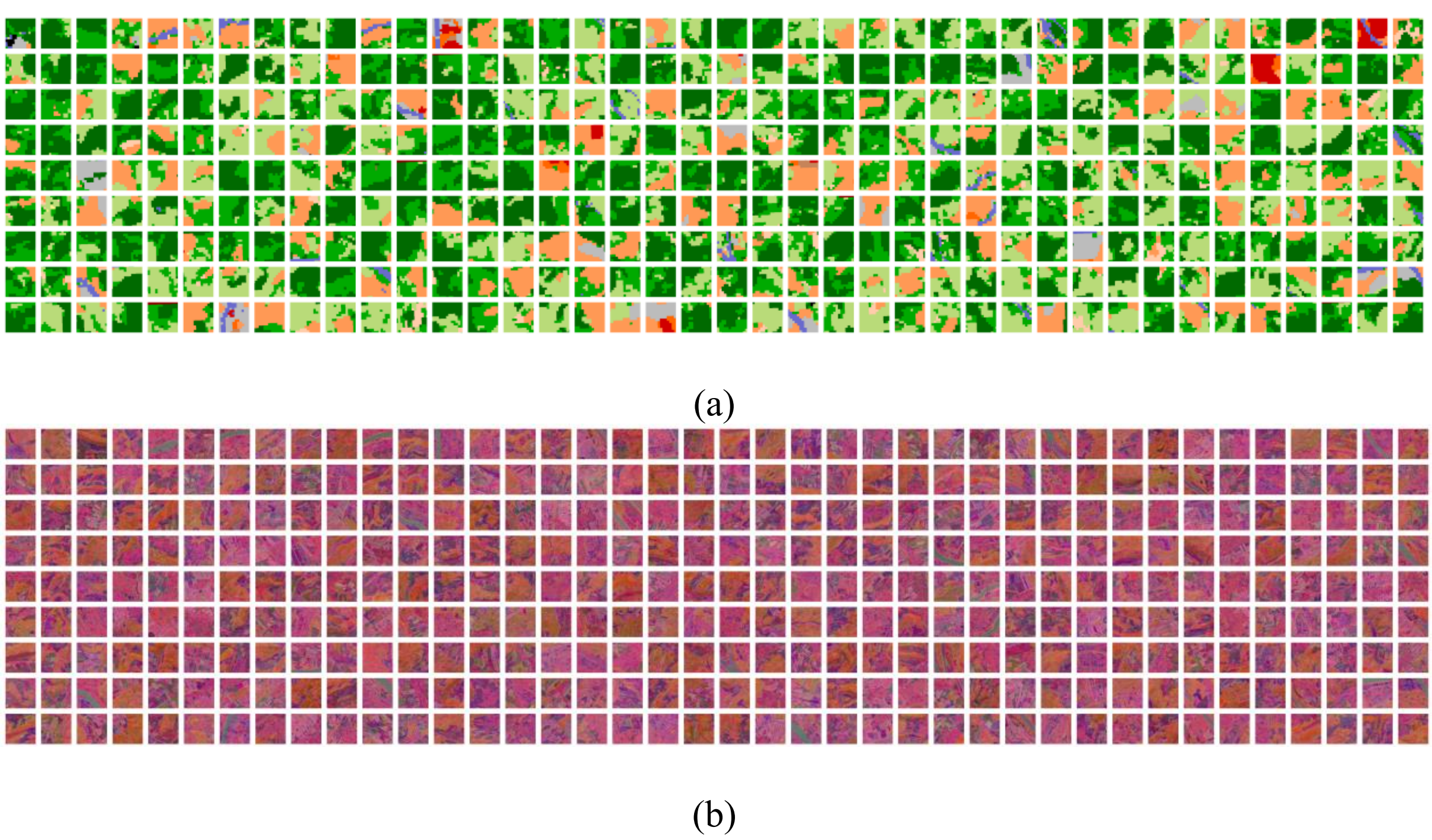

(a)

(b)

**Figure 5:** Illustration of training data for the city of Basel. (a) LCZ labels, and (b) corresponding TESSERA 2024 image patches. All patches have the same spatial extend of 128 × 128 pixels. For TESSERA, only the first three embedding dimensions out of 128 are shown as RGB. Similar datasets were produced for all models, input datasets and cities (not shown).

Different sampling strategies, such as non-overlapping/overlapping sliding windows, can produce different training results (Hester et al., 2026). To train the U-Net, uniformly sized patches of 128 × 128 pixel were prepared. **Figure 5** shows exemplary patches created as the training dataset for the U-Net from Basel. In the proposed study, the major challenge is the presence of class imbalance where certain LCZ classes have notably fewer samples than others. Therefore, samples were generated by using a class-balanced sampling for all pairs of image and label datasets (Bae & Ban, 2025). The objective is to generate 40 samples per LCZ class to ensure samples' representation of all existing classes (Aires et al., 2025). Firstly, stratified random sampling method selected 40 random points for each LCZ class in the reference LCZ map. Additionally, a minimum centroid distance constraint of 10 meters was applied between the patches to avoid saving nearly identical overlapping patches from adjacent tiles. Each accepted point was defined as centroid of patches and created 128×128 patches. The created patches were saved as a georeferenced GeoTIFF file, preserving its spatial coordinates and coordinate reference system. This step ensures the predicted outputs can be mapped back to their correct geographic location.

In both experiments we made sure that the different LCZ categories are present in a balanced way in both training, test and validation datasets (Bae & Ban, 2025). To ensure reproducibility, a stratified random splitting with a fixed seed (random seed = 42) was used (**Table 4**).

For Exp I (model trained and tested with LCZ maps of the five major Swiss cities), we used the imagery of all five cities, from which a total of 1270 patches for training (60%), 424 patches (20%) for validation and 424 patches (20%) for testing. From the original 2320 patches, 202 patches were eliminated because of a single dominant LCZ class. **Table 4** summarizes the available training data for the three experiments.

For Exp II (model trained and tested with purely Bern LCZ map), 560 patches of feature sets and corresponding labels were created. From the 560 patches, 48 were discarded as more than 80% of pixels were from a single LCZ. The valid patches were split into 307 samples for training (60%), 102 samples (20%) for validation and early stopping, and the remaining 103 patches (20%) for testing.

***Table 4*** *Main characteristics of the three experiments*

| | **Experiments** | | |
|---|---|---|---|
| | **Experiment I** | **Experiment II** | **Experiment III** |
| Cities analyzed | Basel, Bern, Geneva, Lauzanne, Zurich | Bern | Basel, Bern, Geneva, Lauzanne, Zurich |
| Number of cities analyzed | 5 | 1 | 5 |
| Reference LCZ Dataset | WUDAPT<br>Details in Table 2a | Bern LCZ<br>Details in Table 2b | WUDAPT<br>Details in Table 2a |

| Number of LCZ classes present | 13 | 14 | 13 |
|---|---|---|---|
| Original number of image patches (128 x 128) | 2320 | 560 | - |
| Total number of image patches (128 x 128) after eliminating single class patches | 2118 | 512 | - |
| Train size (60%) | 1270 | 307 | n.a. |
| Test size (20%) | 424 | 102 | - |
| Validation size (20%) | 424 | 103 | n.a. |

## 4.2. U-Net classifier

The deep learning model applied in this study is an attention-based U-Net, which excels at land use land cover classification and instance segmentation (Deressu et al., 2025; Hester et al., 2026). The advantage of attention mechanisms provides localization to the spatial regions of interest in the feature maps (Hester et al., 2026; Oktay et al., 2018). The attention-based U-Net model is applied to run with multi-channel input datasets. Compared to other U-Net models, this model shows often superior segmentation quality with less training time (Oktay et al., 2018).

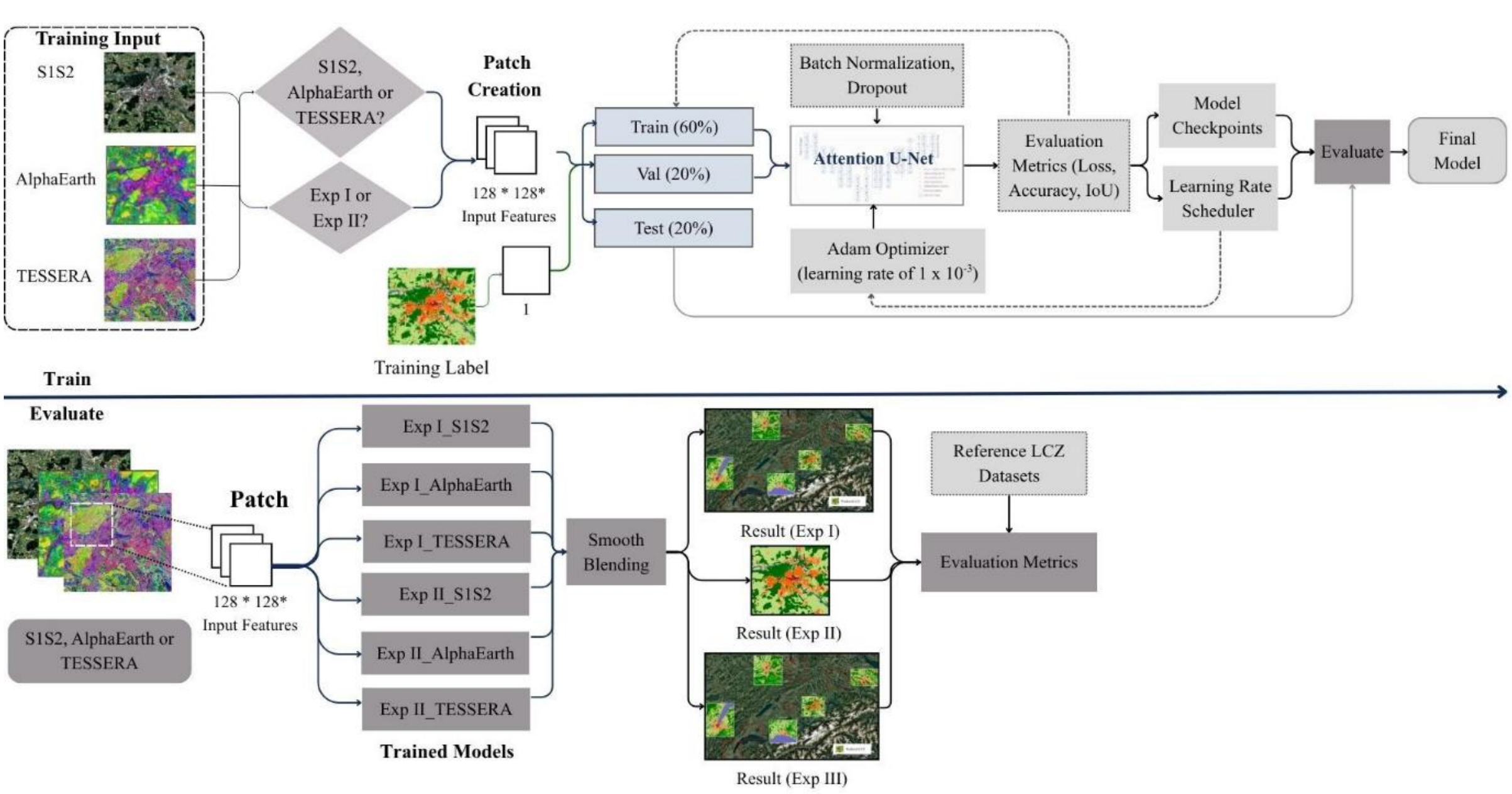

**Figure 6**: Illustration of model training and evaluation. Also shown are details with respect to the chosen U-Net. In the upper part the training step is illustrated. The lower part shows the inference and statistical evaluation.

The attention U-Net architecture is chosen for this study because it is modular and valuable for various application types (Oktay et al., 2018). Importantly, attention U-Nets can be trained with relatively small, labeled datasets while still providing reliable results (Oktay et al., 2018). The preprocessing pipeline set up for this study is inspired by (Ramadhan Ramadhan, 2024) and the training pipeline is inspired by (Bhattiprolu, 2020). The combined pipeline is designed by integrating these inspirations and Python TensorFlow (Abadi et al., 2016) for building deep learning models. The main important components are illustrated in **Figure 6**.

Class imbalance is present across the 17 LCZ categories, with natural land-cover classes such as *Dense trees* (LCZ A) and *Low plants* (LCZ D), accounting for a disproportionate large share of labelled pixels relative to rare urban types such as *Compact high-rise* (LCZ 1) and *Heavy industry* (LCZ 10). To address this, inverse frequency class weights were computed from the training label distribution and incorporated during model training into the composite Dice-Focal loss function (Ahamed & Rahmim, 2023; Azad et al., 2023; Lin et al., 2018).

Different attention U-Net models were developed for each input dataset due to their different features (S1S2: 14 and 16 features for Exp I and Exp II, respectively; AlphaEarth: 64 features; TESSERA: 128 features). However, apart from the input head configuration, each model was fine-tuned using the same configuration such as Adam optimizer, learning rate of 1 x $10^{-3}$, batch size of 16, and Dice + Focal loss function. Evaluation was again performed using 20% of samples reserved for test purposes, and the performance of each model was assessed using each model's IoU and loss scores. Architectural details are shown in **Table 5**.

***Table 5*** *Architecture of the attention U-Net model used for LCZ segmentation*

| Layer | Functions | # Conv Blocks | Activation | Filter Range |
|---|---|---|---|---|
| Input | | - | - | 14/16/64/128 |
| Encoder blocks | Conv2D 3×3<br>BatchNorm<br>Dropout<br>Conv2D 3×3<br>BatchNorm<br>MaxPooling 2×2 | 4 | ReLU | 32→64→128→256→512 |
| Bottleneck (bridge) | Conv2D 3×3<br>BatchNorm<br>Dropout<br>Conv2D 3×3<br>BatchNorm | 1 | ReLU | 512→256 |
| Decoder (expansion) | Conv2DTranspose 2×2<br>Attention Gate | 4 | Sigmoid + ReLU | 256→128→64→32 |

| | Concatenate<br>Conv2D 3×3<br>BatchNorm<br>Dropout<br>Conv2D 3×3<br>BatchNorm | | | |
|---|---|---|---|---|
| Output layer | Conv2D 1×1 | 1 | Softmax | Exp I: 14 (LCZ classes)<br>Exp II: 13 (LCZ classes) |

### 4.3. Model Training and Evaluation

The training process is implemented by using Python TensorFlow, Segmentation Model, Matplotlib and Numpy. Model training was conducted on UBELIX NVIDIA RTX 4090 GPUs and 90GB of Ram (University of Bern, 2026) and testing was performed on a local system equipped with AMD Ryzen 9 3900X 12-Core Processor (without GPU) and 64 GB of RAM.

The batch size was set to 16 during training, and the number of training iterations was 200 epochs. Each model was compiled with Adam Optimizer (learning rate of 1 x $10^{-3}$), integrated with three mechanisms, using total loss function where loss is computed by summing dice and focal loss (**Equation 6**). Focal loss is added to reduce the overconfident on the train dataset.

The different LCZ segmentations were assessed using a combination of segmentation, classification, and spatial statistical measures: IoU, accuracy and loss.

#### 4.3.1. Integration of Early Stopping and Callback Mechanisms

To prevent overfitting and reduce training time, three callback mechanisms were applied during training. First, an *early stopping* callback was applied to monitor the validation IoU at the end of each epoch (Eq 3). If no improvement was observed over 20 consecutive epochs, training was automatically stopped. For completeness, the number of epochs until early stopping is reported when presenting the results.

The *ReduceLROnPlateau* scheduler was used as the second mechanism to monitor validation loss at every epoch. If no improvement was observed over 10 consecutive epochs, the learning rate was automatically reduced by half, and the minimum learning rate is set as $1\times10^{-6}$. This allows the model to learn with finer convergence.

Finally, a model checkpoint callback was used to save the best-performing model weights based on validation IoU. This step ensures that the final model used for evaluation is the best state reached during training rather than the last epoch (Venkatachalam et al., 2025).

#### 4.3.2. Evaluation Parameters

Model performance was reported using Intersection over Union (IoU), accuracy, and loss.
The following equations are applied for the evaluation metrics (Hester et al., 2026):

$$Dice\ Coefficient = \frac{2 \cdot |A \cap B|}{|A| + |B|} \quad (1)$$

Where $|A|$ and $|B|$ are the size of training and validation datasets, respectively.

$$Accuracy = \frac{TP+TN}{TP+TN+FP+FN} \quad (2)$$

$$IoU = \frac{TP}{TP+FP+FN} \quad (3)$$

$$Dice\ Loss\ = 1 - \ Dice\ Coefficient \quad (4)$$

where $TP$ stands for True Positive, $TN$ denotes the number of True Negative samples, $FP$ stands for False Positive, and $FN$ denotes False Negative, respectively.

$$\text{Focal } Loss\ = -\sum_{n=1}^{N}(1 - t_n \cdot y_n)^{\Upsilon}\ \log(t_n \cdot y_n) \quad (5)$$

Where ϒ is a non-negative tunable hyperparameter and it is set as 2 in this study following (Lin et al., 2018).

$$Loss\ = Dice\ Loss + \ Focal\ Loss \quad (6)$$

A combined loss function of dice and focal loss is applied to leverage dice loss's sensitivity to the overlap between the predicted and true masks (Deressu et al., 2025) and focal loss's sensitivity to the data imbalance (Ahamed & Rahmim, 2023; Azad et al., 2023; Lin et al., 2018). By applying combined loss function, the study can balance the influence of majority of the label class. When presenting the results, a clear distinction is made between train and test sets.

### 4.4. Smooth Blending

As the trained U-Net model works on small patches rather than the whole image at once, this creates a problem at the edges of each patch, where the output must generate a whole LCZ image. When these patches are simply placed next to each other without any blending, the final map can look uneven and jagged, with visible boundary lines between patches (Guillaume Chevalier, 2017).

To address this, a smooth blending function based on a second-order spline window was applied when merging overlapping image patches during prediction. Instead of simply joining patches together, this approach assigns lower importance to pixels near the edges of each patch and higher importance to pixels near the center, where predictions are most reliable (Guillaume Chevalier, 2017).

## 5. Results & Discussion

### 5.1. Training across all cities using coarse resolution reference information

Training a single U-Net across five cities using three different sets of 10m input data resulted overall in similar looking maps, with a more fine-grained output compared to the reference data at 100m (***Figure 6***7). In this experiment, a single model was trained and evaluated across five cities using 100 m resolution reference information (WUDAPT LCZ) (for details see **Table 2a**).

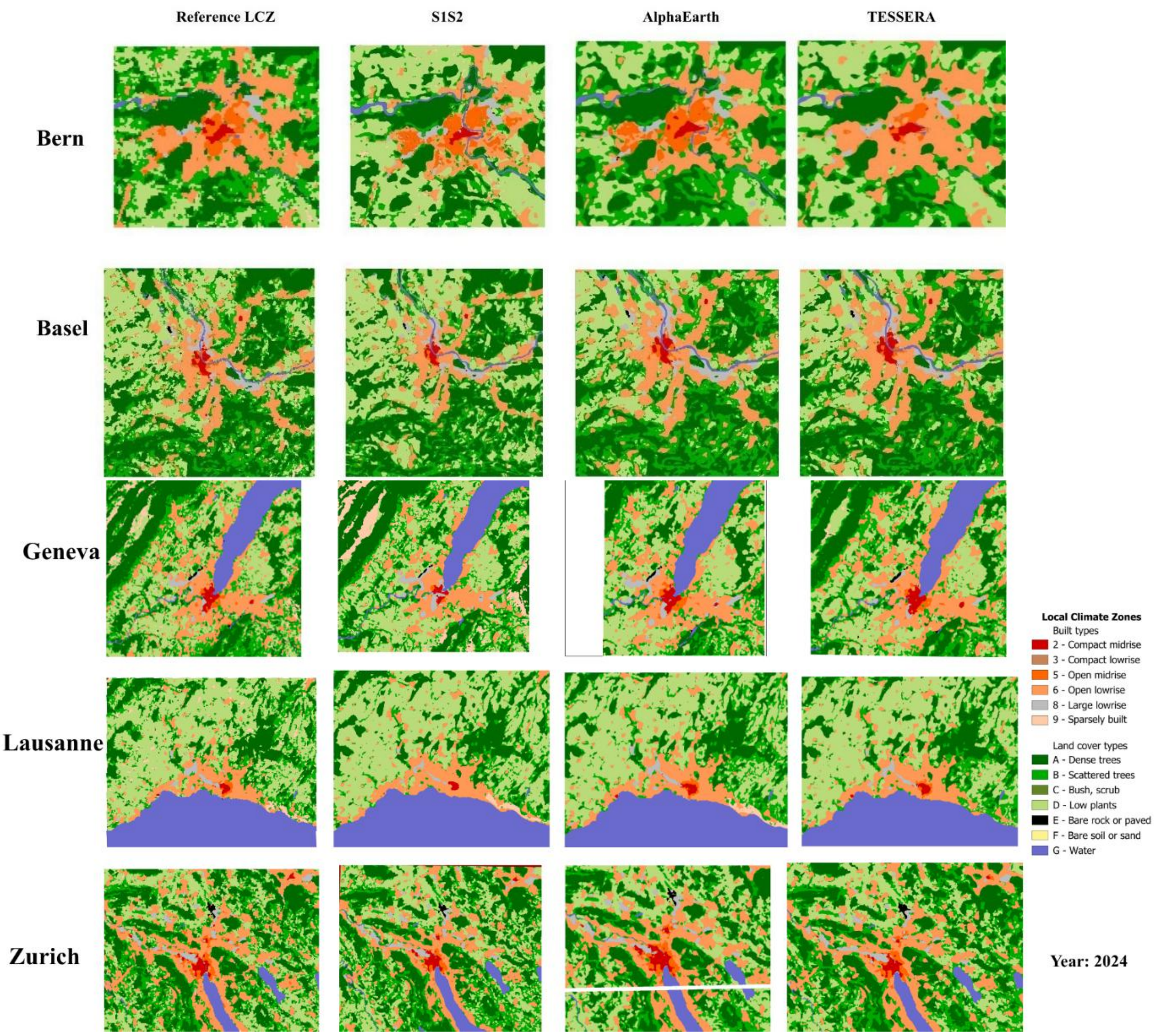

**Figure 7**: Reference and satellite-derived LCZ maps of the five largest cities in Switzerland in 2024 (Experiment I). Small parts of AlphaEarth images are missing in Geneva and Zurich regions – these regions are left blank (white). Note that the reference WUDAPT LCZ dataset is 100 m resolution with only 13 of the 17 classes available for training (see **Table 2a**).

To quantitatively evaluate model performance across the different city structures, results from Experiment I are presented in **Table 6**. Among the three models, TESSERA achieved the best test results with a test IoU of 0.69 and accuracy of 0.80, closely followed by S1S2 and AlphaEarth.

For comparison, **Table 8** reports the results obtained for the city of Bern, using higher resolution (78 m) LCZ reference information. Compared to Experiment II (**Table 7**), all three models show a noticeable drop in test performance because of the added complexity of training across multiple cities and the use of coarser resolution (100 m WUDAPT) reference data. The relative ranking between input datasets is largely preserved, with TESSERA maintaining its advantage.

All models, and TESSERA in particular, show a pronounced gap between training and test performance (train vs test ΔIoU of approximately 0.15 units), suggesting some overfitting to the peculiarities of the multi-city training set. S1S2 shows a comparably smaller train-test gap, indicating a more stable generalization across cities despite its lower absolute test performance.

***Table 6*** *Comparison of LCZ segmentation performance in Exp I when a single model was trained across all cities/ datasets and evaluated across five major Swiss cities. In* ***bold*** *the best accuracies and IoU for the test set.*

| | **S1S2** | **AlphaEarth** | **TESSERA** |
|---|---|---|---|
| Architecture | | Attention U-Net<br>Details in Table 5 | |
| Predictor set | | Details in Table 3 | |
| Input Channels | 16 | 64 | 128 |
| Patch Size | | 128 × 128 px | |
| | | | |
| Spatial Resolution of predictor variables | | 10 m | |
| No. of Classes | | 14<br>Details in Tables 2a and 4 | |
| Spatial resolution of LCZ reference | | 100 m | |
| No. of epochs until early stopping | | 170 | |
| Best epoch | | 148 | |
| Train Loss | | 0.13 | |
| Test Loss | | 0.33 | |
| Train Accuracy | | 0.89 | |
| Test Accuracy | | 0.79 | |
| Train IoU | | 0.81 | |
| Test IoU | | 0.67 | |
| Loss Function | | Dice + Focal | |
| Optimizer | | Adam (lr=$1\times10^{-3}$) | |
| Study Area | | Basel, Bern, Geneva, Lausanne, Zurich | |
| Reference Labels | | WUDAPT LCZ map (100 m)<br>Details in Table 2a | |

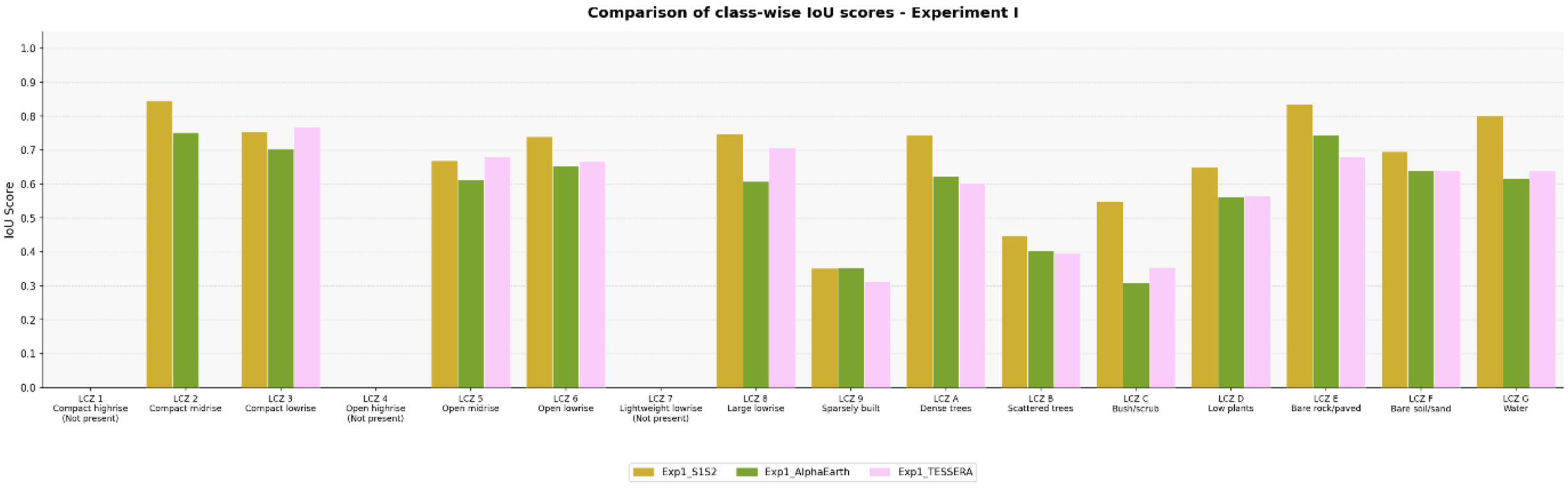

***Figure 8**: Comparison between class-wise IoU scores of Experiments I. Please note that LCZ classes 1, 4, 7 and 10 are not present in the reference dataset.*

The class-wise IoU scores shown in **Figure 8** reveal clear differences in how well the three models identify each LCZ type in the multi-city Experiment I. All three models perform consistently well on built-up classes such as LCZ 2 (*Compact midrise*), 5 (*Open midrise*), 8 (*Large lowrise*), and on natural classes such as LCZ A (*Dense trees*), D (*Low plants*), E (*Bare rock or paved*), and G (*Water*), where IoU scores stay around or above 0.80 for most models.

The weakest classes across all three models are LCZ 9 (*Sparsely built*), B (*Scattered trees*), and C (*Bush and scrub*), with IoU values around 0.55-0.70. One thing worth to mention is all the models failed to detect LCZ 3 (*Compact low-rise*) and LCZ F (*Bare soil / sand*). These classes are severely underrepresented in the training data; and they frequently co-exist within a single pixel footprint, leading to ambiguous class assignment and reduced label consistency during both training and evaluation.

Differences between the three models are generally small but follow a consistent pattern. AlphaEarth and TESSERA achieve very similar scores across all classes. S1S2 remains competitive on the denser built-up classes (LCZ 2, 6, 8) but underperforms the two embedding-based models on LCZ 5 (*Open midrise*), 9 (*Sparsely built*), A (*Dense trees*), E (*Bare rock or paved*), and G (*Water*). This indicates that the embedding-based representations encode richer spatial and temporal information that helps for sparse-built and natural classes under multi-city training conditions.

The relatively small spread between models on most classes also suggests that under the Exp I setting the choice of input representation matters less than the inherent class separability and reference data quality. Despite the scarcity of training examples for rare LCZ types, all three models develop sufficient representational capacity to identify the most common urban and natural classes.

## 5.2. Training with city-specific reference data with higher spatial resolution (Bern LCZ)

Exemplary results of the prediction model compared to the ground truth in the evaluation set for the 14-class LCZ map of Experiment II (Bern) are shown in **Figure *9***. Compared to the original reference data at 78 m, all models lead to more fine-grained maps, better reflecting the local landscape, respectively, urban structure of Bern.

The overview with zoom-ins in **Error! Reference source not found.Figure *10*** (b) exemplifies that embeddings often improve the LCZ mapping performance compared to simple annual composites. In the highlighted example, we do not only find the urban structure better resolved, but we also find the river was smoothed out (blue) around the main forest patch north of the city in the LCZ map generated by S1S2-based model. However, we also noticed one vegetated area which S1S2 and AlphaEarth misclassified as *Water* (LCZ G) while TESSERA correctly classified it as *Low plants* (LCZ D). This suggests that TESSERA's richer temporal encoding preserves the subtle spectral and temporal distinctions that single-composite and patch-based representations can confuse.

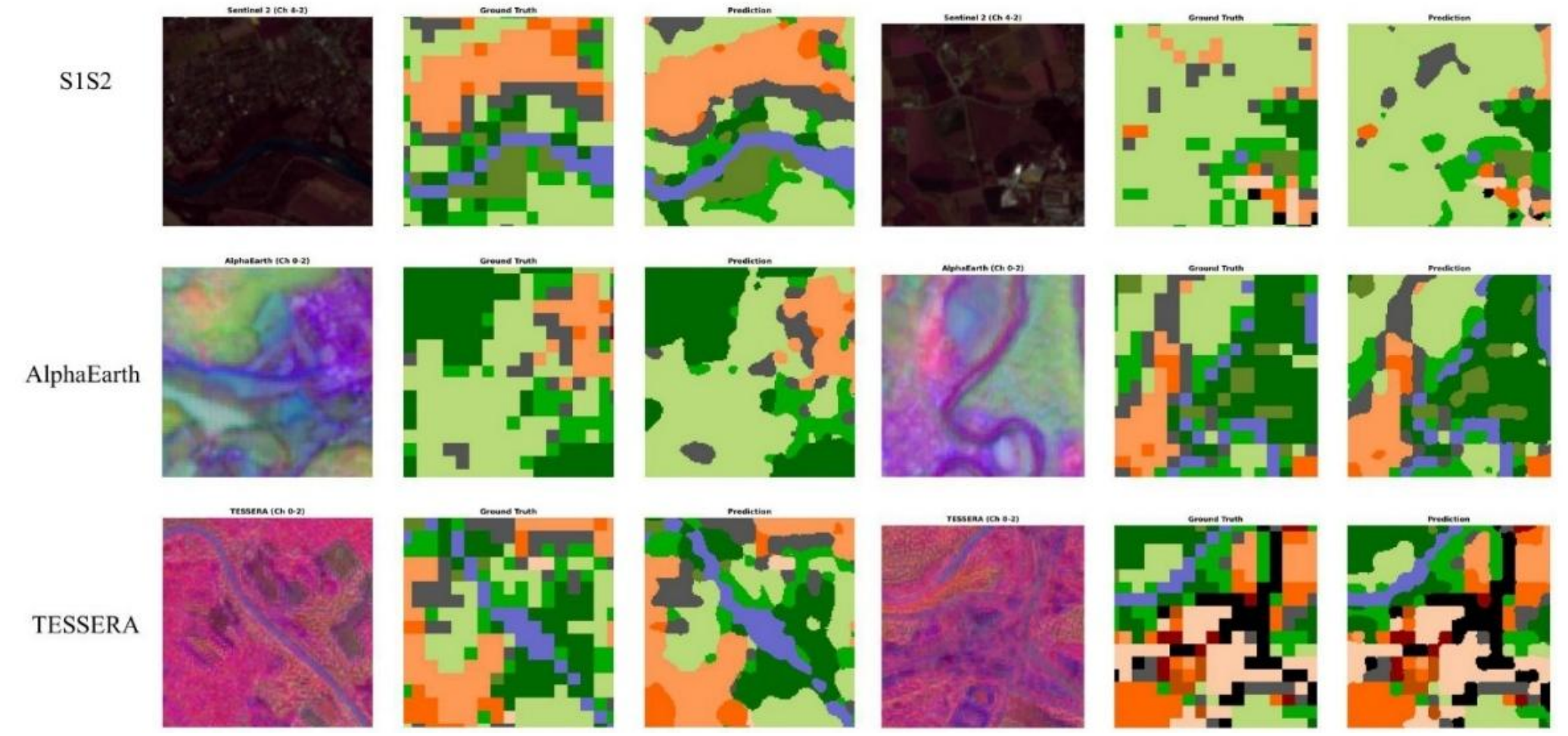

**Figure 9**: Exemplary classification results with examples from Experiments II (Bern) together with inputs and reference labels (“ground truth”). The ground truth in this experiment is provided at 78m.

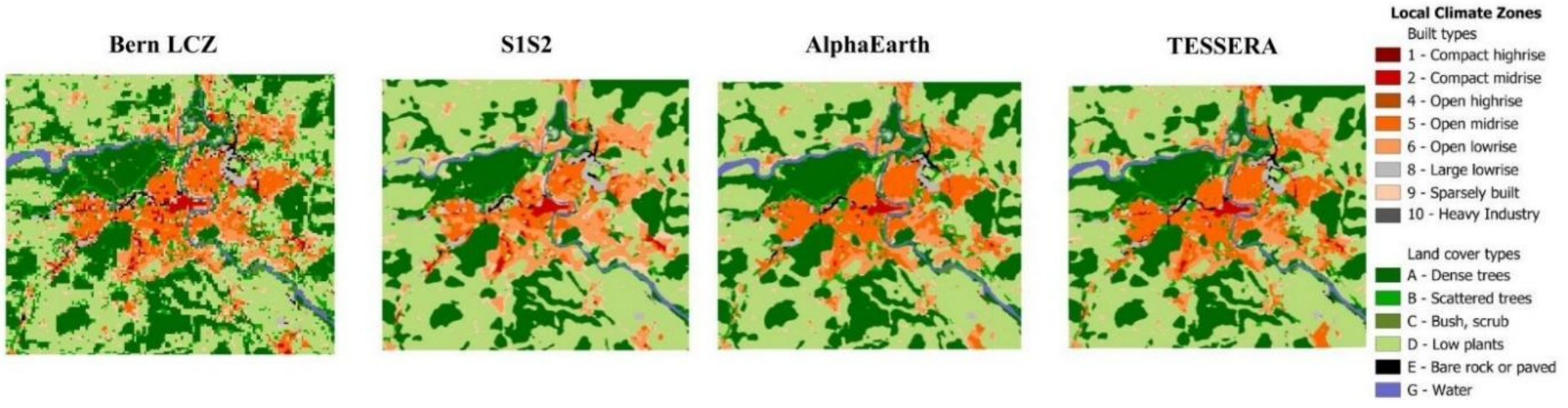

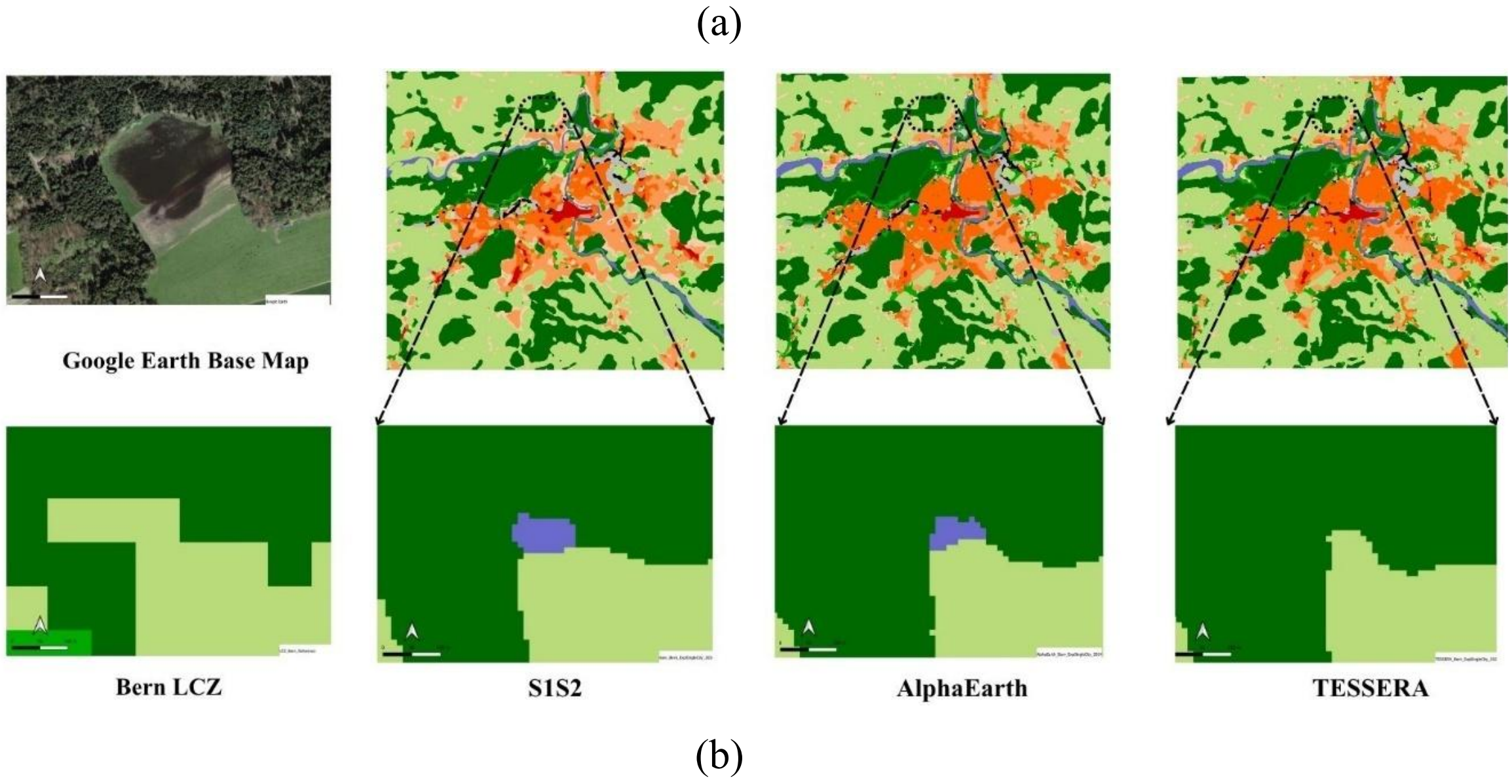

In quantitative terms (**Table 78**), the performance of the different models is evaluated by the

**Figure 10**: LCZ maps of Bern (Experiment II). Shown are both the reference dataset as well as satellite-derived maps for the study year 2024. (a) Comparison of LCZ generated by different input datasets compared to the reference data; (b) zoomed-in areas with some larger discrepancies found between the modelled maps. In the chosen example S1S2 and AlphaEarth misclassified the vegetation field as Water (LCZ G), while TESSERA correctly classified it as Low plants (LCZ D).

IoU (Intersection over Union). Using this statistic all models perform similarly well, TESSERA yielding slightly higher test IoU compared to the annual S1S2 composites and AlphaEarth.

***Table** 7 Comparison of LCZ segmentation performance in Bern, Switzerland (Experiment II) using three different input datasets: S1S2, AlphaEarth and TESSERA. Details referring to the different input datasets are given in **Table 3**. In **bold** the best results based on the independent test dataset.*

| | **S1S2** | **AlphaEarth** | **TESSERA** |
|---|---|---|---|
| **Architecture** | | Attention U-Net<br>Details in Table 5 | |
| **Predictor set** | | Details in Table 3 | |
| **Input Channels** | 16 | 64 | 128 |
| **Patch Size** | | 128 × 128 px | |
| **Spatial Resolution of predictor** | | 10 m | |

| variables | | | |
|---|---|---|---|
| **No. of Classes** | | 14<br>Details in Tables 2 and 4 | |
| **Spatial resolution of LCZ reference** | | 78 m | |
| **No. of epochs until early stopping** | 124 | 196 | 177 |
| **Best epoch** | 108 | 193 | 148 |
| **Train Loss** | 0.11 | 0.07 | 0.08 |
| **Test Loss** | 0.10 | 0.06 | 0.07 |
| **Train Accuracy** | 0.93 | 0.95 | 0.95 |
| **Test Accuracy** | 0.87 | 0.89 | **0.90** |
| **Train IoU** | 0.87 | 0.91 | 0.90 |
| **Test IoU** | 0.77 | 0.81 | **0.82** |
| **Loss Function** | | Dice + Focal | |
| **Optimizer** | | Adam (lr=$1\times10^{-3}$) | |
| **Study Area** | | Bern, Switzerland | |
| **Reference Labels** | | Bern LCZ (78 m)<br>Details in Table 2 | |

All three models achieved high test accuracy and IoU scores exceeding 0.87 and 0.77, respectively. The TESSERA-based model achieved the highest test IoU score followed by AlphaEarth (0.81) and the model using annual composites. The difference between AlphaEarth and TESSERA is negligible, but both clearly outperform the S1S2 composites in term of test IoU.

Interestingly, the conventional annual S1S2 (Sentinel-1 and Sentinel-2) composites which also incorporate the high resolution Swisstopo DSM and DTM (**Table 4**), can achieve the high evaluation scores of the embeddings in the training set. However, this did not carry over to the validation set, where TESSERA and AlphaEarth better generalized slightly better. This suggests that despite the added elevation detail, the S1S2 composite model learned patterns too specific to the training data, while the embedding-based models generalized more reliably to unseen patches.

Most models from both Exp I and II perform consistently well on common urban classes such as LCZ 2, 5, 6, and 8, where IoU scores stay above 0.70 for most models across both experiments (**Figure 11**). Across all models and both experiments, *LCZ 9 (Sparsely built), B*

*(Scattered trees)*, and *C (Bush and scrub)* stand out as the weakest classes. Two compounding factors explain this: first, *LCZ 9 (Sparsely built) and C (Bush and scrub)* classes are severely underrepresented in the training data; and second, they frequently co-exist within a single pixel footprint. This leads to ambiguous class assignments and reducing label consistency during both training and evaluation.

**Comparison of Classwise IoU Scores**

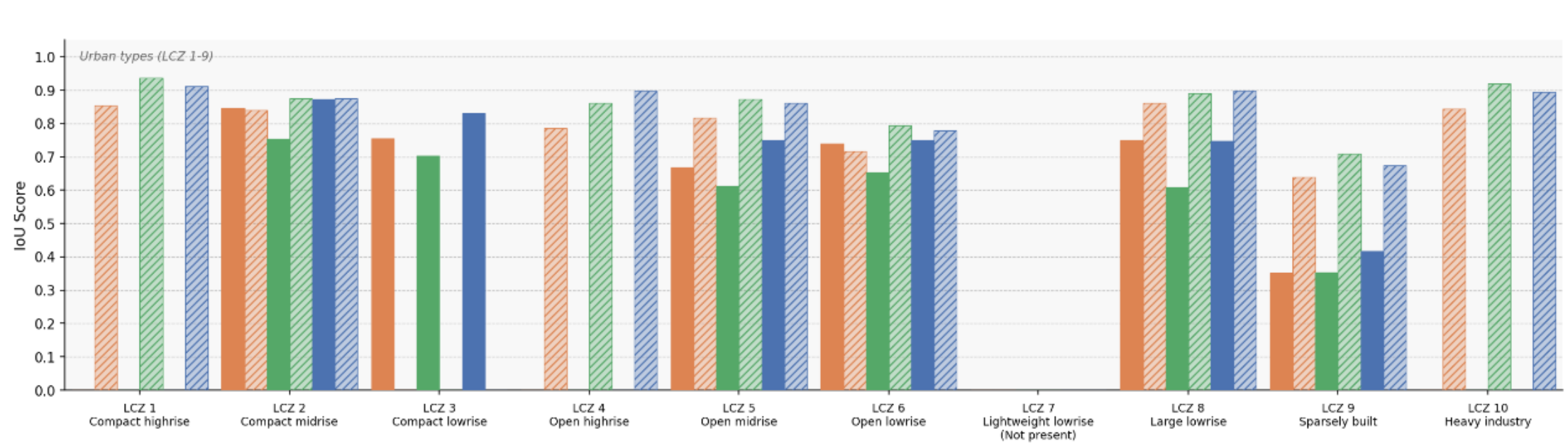

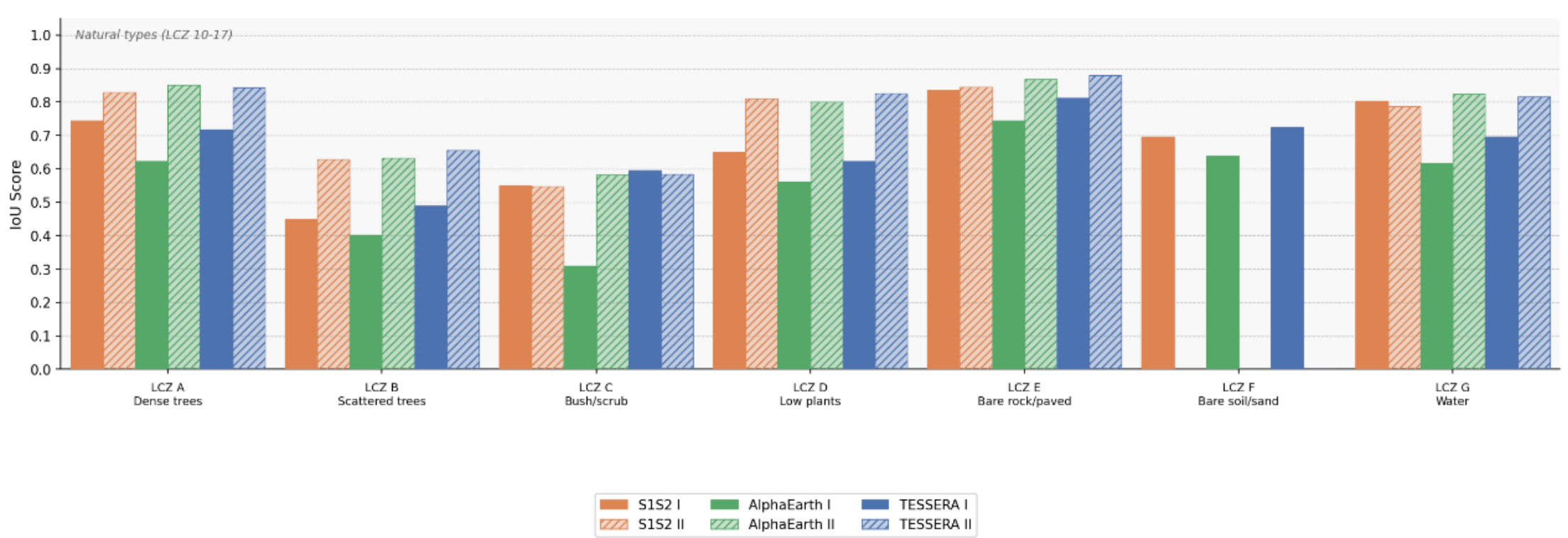

***Figure 11**: Comparison between class-wise IoU scores of Experiments I and II. Please note that for Experiment I LCZ classes 1, 4, 7 and 10 are not present in the reference dataset. For Experiment II, LCZ classes 1, 3 and F are missing. Note that results for Exp II only refer to the city of Bern, whereas to the five Swiss cities in Exp I.*

As briefly mentioned, looking across the two experiments, Exp II models perform better in most classes compared to Exp I. TESSERA achieves stronger results on most urban LCZ types, particularly on sparse classes such as LCZ 1 (*Compact highrise*) and 3 (*Compact lowrise*), suggesting that its spectral signatures capture certain urban morphologies more effectively under multi-city training conditions. S1S2 of Experiment I remains competitive on denser urban classes such as *LCZ 2 (Compact midrise), 6 (Open lowrise), and 8 (Large lowrise)*. On the other hand, AlphaEarth noticeably underperforms on natural LCZ types in both experiments, with particularly low scores for *LCZ B (Scattered trees) and C (Bush, scrub)*, while remaining competitive on built-up classes such as LCZ 8 (*Large lowrise*) in Experiment

II. In Experiment I, S1S2 outperforms the embedding-based models on *LCZ A (Dense trees), LCZ E (Bare rocks or paved) and LCZ G (Water)*, suggesting that S1S2 spectral time-series retain an advantage for certain vegetation, bare and water surface types.

The results clearly reflect the combined challenges of strong class imbalance in the training data, limited spatial extent of these categories in the study area, and the difficulty of learning robust discriminative features for underrepresented classes in the spectral and embedding space. Despite the scarcity of training examples for rare urban morphologies, the embedding-based models still develop some representational capacity to identify them.

**Table 8**: Per-class reference and estimated areas (in $km^2$) of Bern (Experiment II) with a total of 14 LCZ classes. The spatial resolution of the reference LCZ is 78 m, whereas the satellite-derived maps have a spatial resolution of 10m. Under-estimations are color-coded as brown and over-estimations as green. For each class, the best result is shown in a lighter color with **bold** acreage.

| Class label | Class definition | Reference LCZ map (in $km^2$) | Modeling results (in $km^2$) | | |
|---|---|---|---|---|---|
| | | | S1S2 | AlphaEarth | TESSERA |
| LCZ 1 | Compact high-rise | 0.07 | 0.00 | 0.00 | 0.00 |
| LCZ 2 | Compact mid-rise | 2.29 | 0.61 | 1.09 | **1.45** |
| LCZ 4 | Open high-rise | 0.59 | 0.01 | **0.06** | 0.01 |
| LCZ 5 | Open mid-rise | 15.15 | 19.66 | 19.44 | **19.06** |
| LCZ 6 | Open low-rise | 11.56 | 12.85 | 9.83 | **11.46** |
| LCZ 8 | Large low-rise | 2.42 | 1.63 | 2.51 | **2.30** |
| LCZ 9 | Sparsely built | 10.02 | 5.40 | **6.65** | 5.19 |
| LCZ 10 | Heavy industry | 0.17 | 0.00 | 0.00 | 0.00 |
| LCZ A | Dense trees | 38.87 | 42.06 | 42.82 | **41.80** |
| LCZ B | Scattered trees | 14.86 | 6.16 | **7.35** | 6.47 |
| LCZ C | Bush / scrub | 2.10 | 0.07 | 0.09 | **0.12** |
| LCZ D | Low plants | 56.53 | 66.30 | **64.33** | 66.37 |
| LCZ E | Bare rock / paved | 1.33 | 0.52 | **1.13** | 1.07 |
| LCZ G | Water surface | 3.08 | 3.78 | 3.72 | **3.52** |

To assess how well the models capture the actual acreage of the different classes, in **Table 89Error! Reference source not found.**, the predicted LCZ class areas of three models for Bern were compared against the corresponding reference map. Overestimated LCZ classes are highlighted in green while the underestimated LCZ classes are shown in brown. For each class, the best results are shown in a lighter color with bold numbers.

The resulting pattern is relatively inconsistent with overall relatively large differences for all models, also due to the mismatch between the resolutions of reference map (78m) and the predicted maps (10m). The resolution mis-match also aligns with the fact that classes with low abundance are mostly underestimated in the predicted maps but overestimated for the classes with higher acreage. In this respect, we see almost no difference between the three models. In most cases, the three input datasets are aligned in their over-estimation, respectively, under-estimation. What stems out however, is that none of the 14 classes is best predicted using S1S2, despite the integration of high resolution DTM and DSM datasets(Federal Office of Topography swisstopo, 2024, 2025).

### 5.3. Cross-reference Evaluation

To separate the effects of training label resolution from evaluation reference resolution, we computed another experiment for Bern: each prediction (from the Experiment I model trained on WUDAPT 100m labels and the Experiment II model trained on the 78m Bern reference) was scored against both reference datasets. The results (**Table 10**) reveal a consistent pattern across all three input sources: each model scores highest when evaluated against the reference dataset it was trained on, and substantially lower when evaluated against the other. The observed gap is around 16-22 percent points and arises purely from the choice of reference; the underlying predictions are unchanged. Coarse reference data thus does not merely under-evaluate fine-scale predictions, it imposes its own definition of correctness on the assessment. Holding the evaluation reference fixed at 78m, models trained on the higher-resolution Bern labels outperform their WUDAPT-trained counterparts by approximately 20 percentage points Overall Accuracy, quantifying the value of higher-resolution training data once the evaluation bias is removed.

**Table 9**: Cross-reference evaluation of Bern LCZ predictions of Exp I and II. Each model (S1S2, AlphaEarth, TESSERA) was scored against both the WUDAPT (100 m) and the Bern LCZ (78 m) reference datasets, for predictions generated by both the Exp I model (trained on multi-city 100 m WUDAPT labels) and the Exp II model (trained on the 78 m Bern LCZ labels). Overall accuracy (OAA) and mean IoU are reported.

| Model | Train res | Evaluation res | OAA | Mean_IoU |
|---|---|---|---|---|
| **S1S2** | 100 m | 100 m | 0.73 | 0.44 |
| **S1S2** | 100 m | 78 m | 0.57 | 0.18 |
| **AlphaEarth** | 100 m | 100 m | 0.69 | 0.41 |

| **AlphaEarth** | 100 m | 78 m | 0.54 | 0.22 |
|---|---|---|---|---|
| **TESSERA** | 100 m | 100 m | 0.76 | 0.48 |
| **TESSERA** | 100 m | 78 m | 0.55 | 0.18 |
| **S1S2** | 78 m | 100 m | 0.52 | 0.24 |
| **S1S2** | 78 m | 78 m | 0.75 | 0.35 |
| **AlphaEarth** | 78 m | 100 m | 0.54 | 0.27 |
| **AlphaEarth** | 78 m | 78 m | 0.77 | 0.37 |
| **TESSERA** | 78 m | 100 m | 0.53 | 0.26 |
| **TESSERA** | 78 m | 78 m | 0.77 | 0.38 |

### 5.4. Generalization

To assess the learning behavior of the different models, and to get some insights into the generalization from training to validation data, ***Figure 12*** shows the IoU metric (a) and loss (b) for the 200 epochs of the model training. The curves are shown separately for training and validation datasets and for the two experimental settings. Note that only the IoU metric is used to optimize the model (**Eq 3**) and that the loss (**Eq 6**) is only shown for completeness.

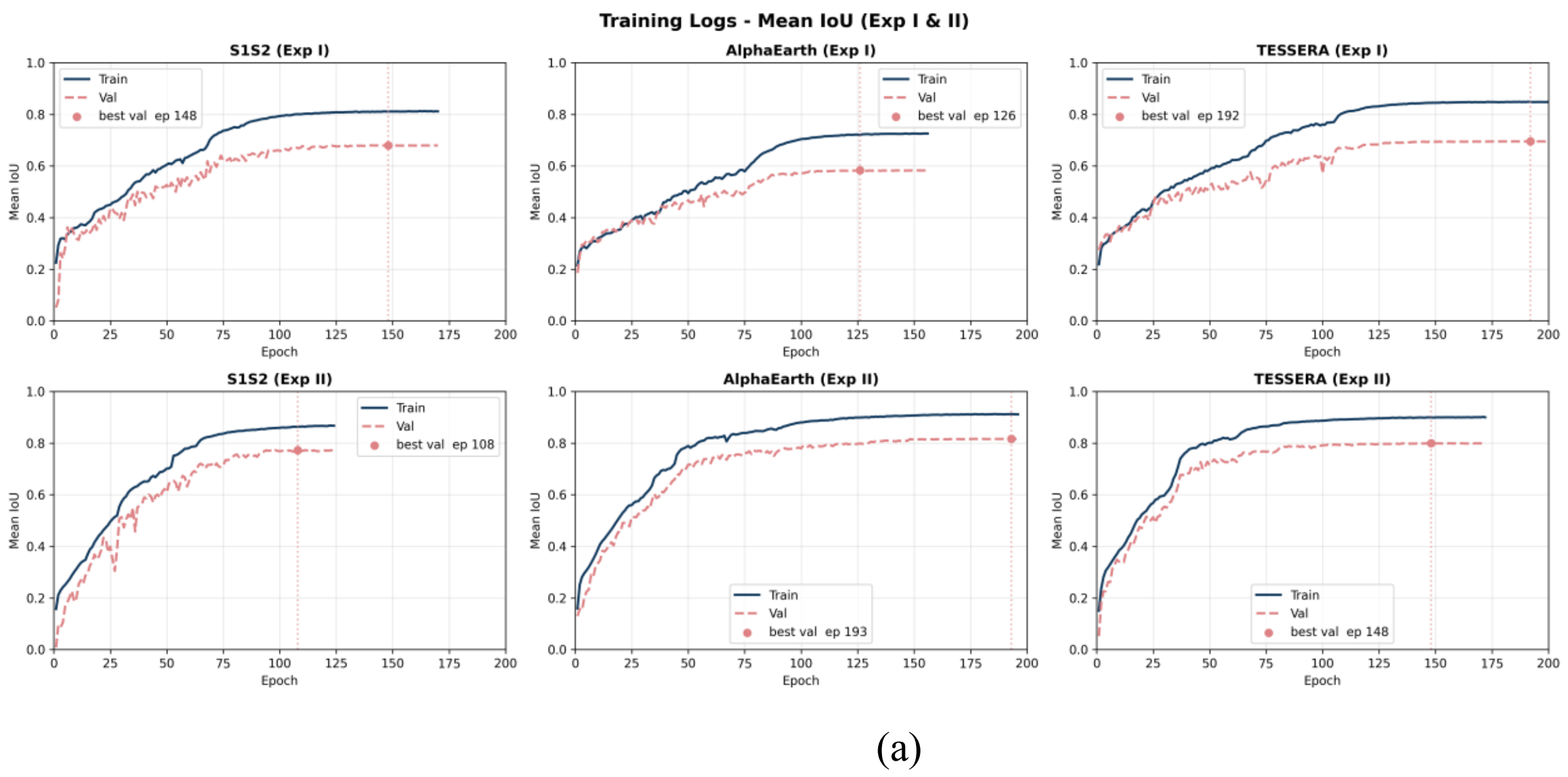

(a)

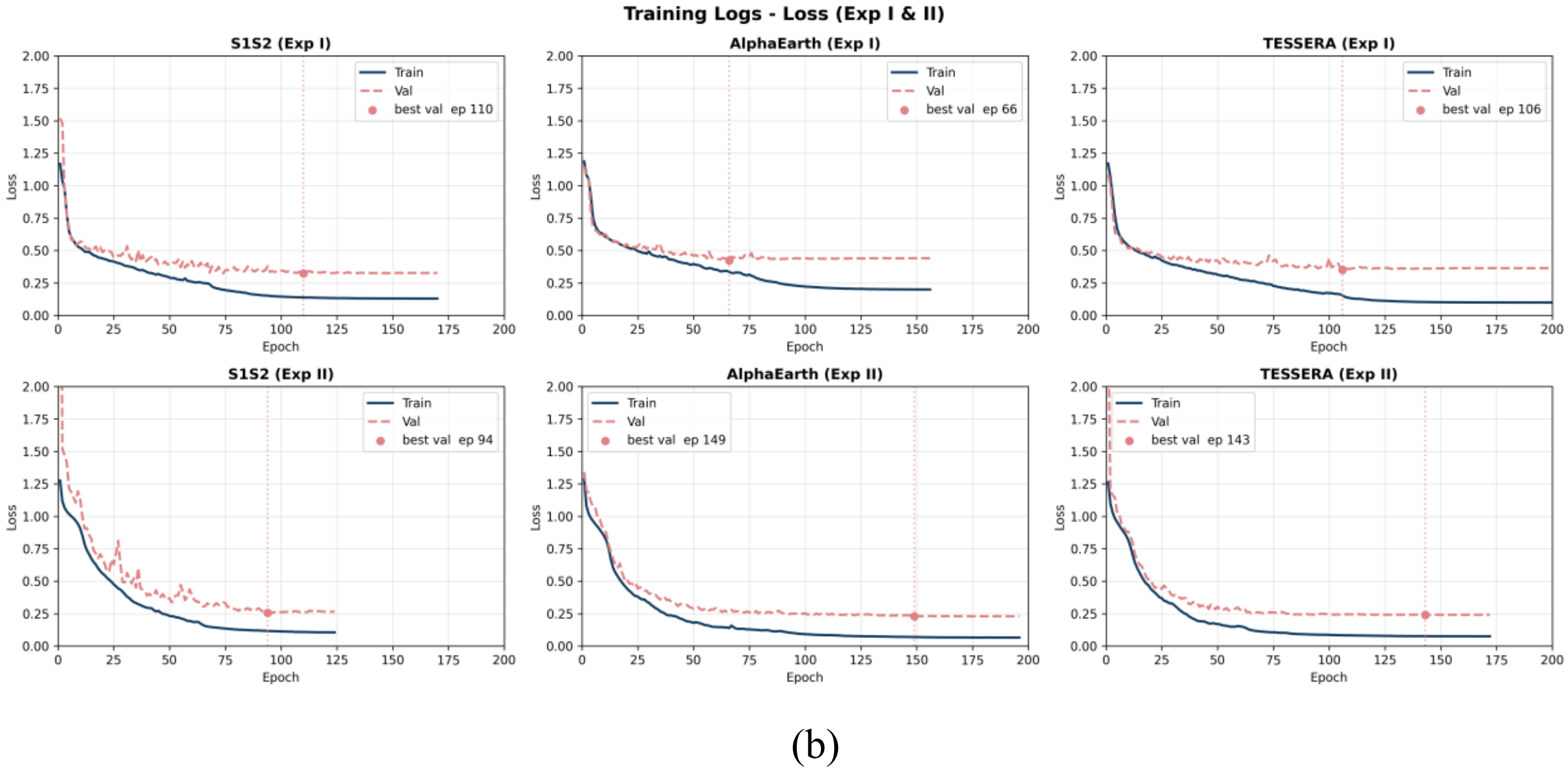

(b)

**Figure 12**: Evolution of training and validation IoU (a) and loss scores (b) during model training providing information with respect to the generalization of models. Note that the loss (Eq 6) is shown here only for completeness – it does not enter in the model optimization

**Figure 12** shows how IoU and loss evolved over the training epochs for all six models in the two experiments. In Exp I (multiple cities), all three models exhibit a substantial gap between training and validation curves, with training IoU continuing to rise while validation performance plateaus around epoch 100. In Exp II (single city), the three models converge smoothly, with training and validation curves tracking closely throughout, indicating good generalization to unseen data. This consistent pattern across all models reflects the difficulty of the multi-city training setting and suggests that the models learn features that do not fully transfer to the held-out test dataset. TESSERA converges most slowly in Exp I but achieves the highest final validation IoU, suggesting that its rich embeddings require more training epochs to fully exploit but ultimately reach comparable generalization.

**Figure 13** provides a direct comparison of validation IoU and loss across all three models within each experiment. In Exp I (**Figure 13** a), the divergence between S1S2 and TESSERA versus AlphaEarth becomes clearly visible after approximately epoch 75, with AlphaEarth stabilizing at a lower plateau. In Exp II (**Figure 13** b), all models converge to similar validation loss values with considerably less noise, confirming that the single-city training setting provides a more stable and generalizable learning signal.

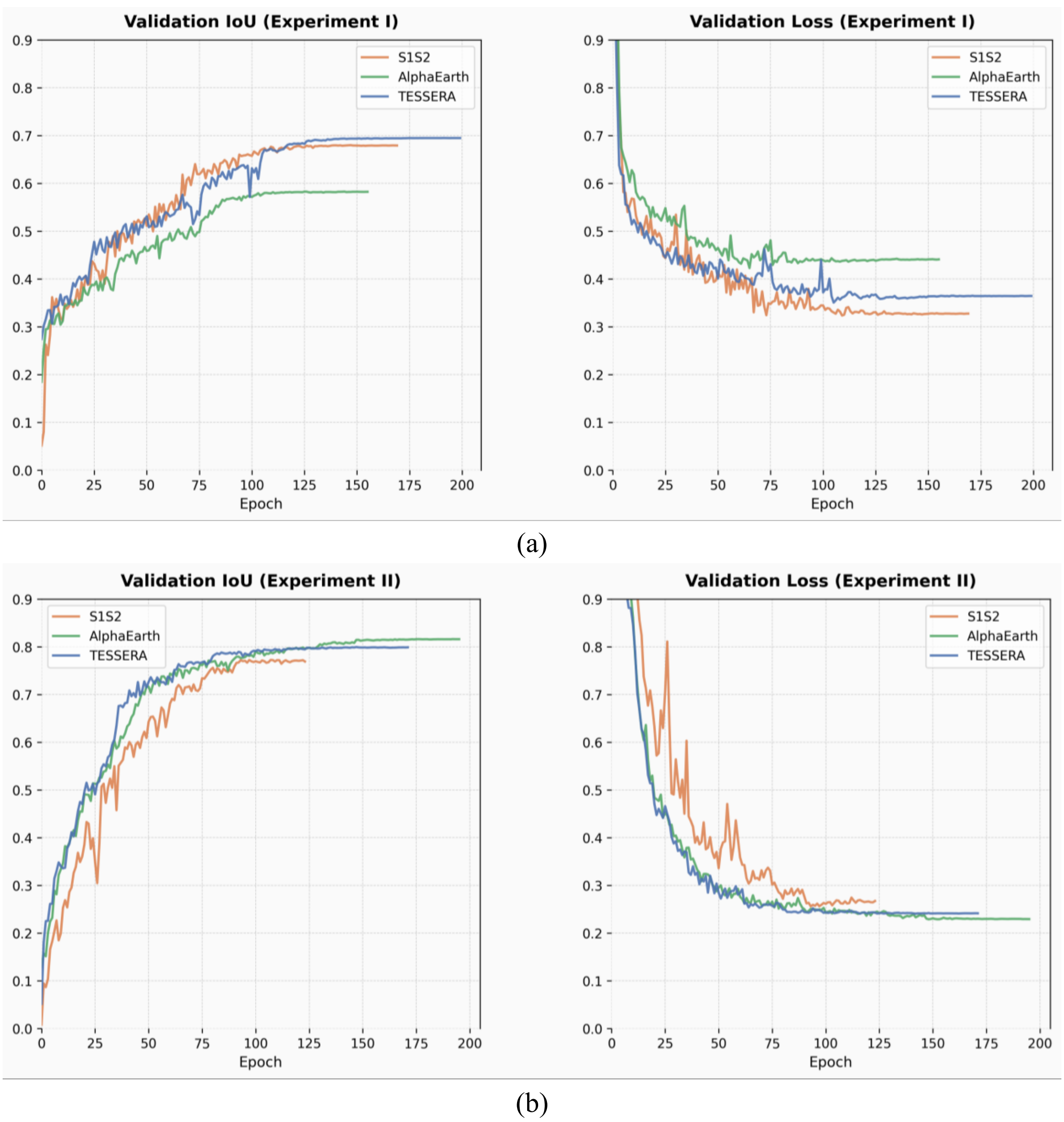

***Figure 13**: Evolution of the Validation IoU and loss during model training for three input sources and two experiments: (a) Exp I, (b) Exp II. All models are trained till 200 epochs and early callback is set up.*

### 6.4.Temporal transferability of developed models

As in all other machine and deep learning applications, the availability of high-quality reference data is the main bottleneck in satellite-based mappings. Hence, it is of utmost importance to develop classifier which can be robustly transferred from one year (e.g., the year the training data was collected) to another year.

This aspect is studied in Exp III, where the pre-trained models of Exp I from 2024 are utilized to predict the LCZ classes using satellite data from 2025. If the LCZ remain relatively unchanged between 2024 and 2025, one would expect almost unchanged LCZ maps.

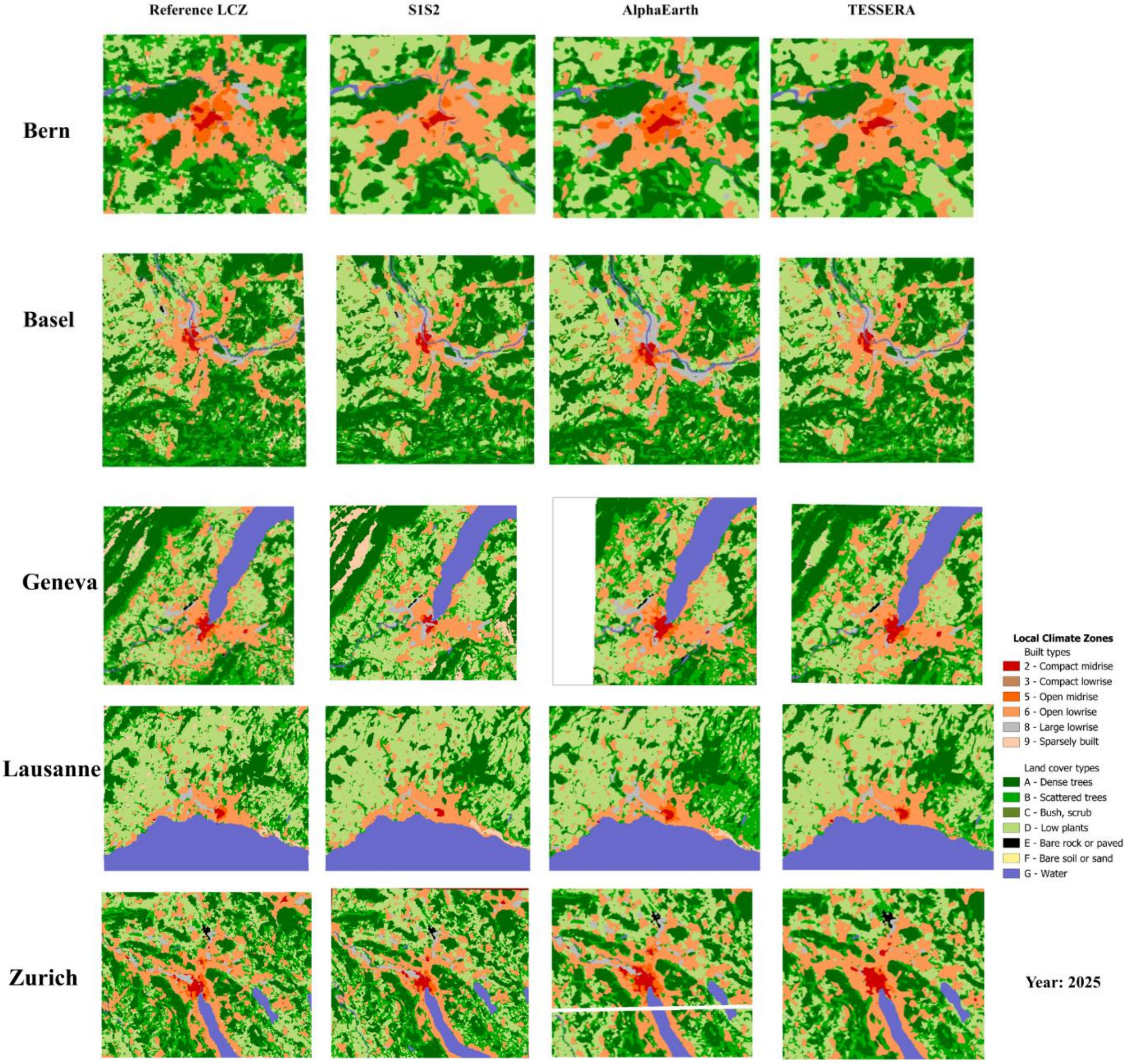

**Figure 14**: Predicted LCZ maps for the five largest cities in Switzerland using models pre-trained on data from 2024 with inputs from 2025. The left column shows for comparison the reference datasets. Only the trained models from Exp I are applied. Note that for 2025, the AlphaEarth datasets for Geneva and Zürich are incomplete.

**Figure 14** shows the resulting 2025 LCZ maps for all five cities together with their reference maps. Please note that AlphaEarth images for Geneva and Zurich regions are presented with small data gap.

Overall, the predicted maps capture the broad spatial distribution of LCZ classes well, and the general urban structure of each city remains clearly recognizable across different inputs. This suggests that the trained models retain sufficient representational capacity to produce meaningful LCZ maps one year beyond their training period. This visual inspection is confirmed through statistical analysis. Indeed, the 2025 predictions do not differ much from the accuracy achieved on same-year result as shown in **Tab 11**. When comparing 2024 and 2025 results, a general decline in accuracy is observed across models and cities, reflecting the challenge of temporal transfer without retraining. S1S2 shows the most stable performance with only slight accuracy differences between the years. TESSERA also transfers reasonably well, with moderate declines in most cities. AlphaEarth exhibits the largest year-to-year variability, with notable drops in Bern (73.9% to 60.1%) and Basel (77.4% to 68.8%), suggesting greater sensitivity to temporal changes in input data. The anomalous accuracy increase in Geneva and Zurich for AlphaEarth should be disregarded given the partial data coverage in both years.

The (seasonal) S1S2 composites thus remove very much of the subtle temporal variations captured by embeddings. This is further confirmed by the class-wise area statistics shown in Annex (Error! Reference source not found.) which also confirm that year-to-year variations in estimated areas are the most stable for S1S2, followed by TESSERA. Together, these results confirm that TESSERA provides the strongest overall accuracy when evaluated against data from the year of model training, while S1S2 demonstrates the most consistent temporal transferability across years.

***Table 10*** *Summary of cross-year accuracy assessment. The model trained from Exp I was applied to data from 2024 and 2025, respectively, to assess model transferability across time. In bold the best transfer results for 2025 (e.g. Exp III). Please note that study areas with partial data gaps in AlphaEarth embeddings are presented with * and italic font because part of the data is missing, those results are not included in the model ranking.*

| City | Year | S1S2 | AlphaEarth | TESSERA |
|---|---|---|---|---|
| Basel | 2024 | 77.0 | 77.4 | 80.0 |
| Basel | 2025 | **76.2** | 68.8 | 75.9 |
| Bern | 2024 | 72.0 | 73.9 | 74.9 |
| Bern | 2025 | **70.0** | 60.1 | 67.6 |
| Geneva | 2024 | 77.1 | *63.7** | 80.6 |
| Geneva | 2025 | 75.3 | *79.0** | **78.1** |
| Lausanne | 2024 | 83.3 | 82.7 | 84.0 |
| Lausanne | 2025 | **82.6** | 81.6 | 82.3 |
| Zurich | 2024 | 75.4 | *34.2** | 78.4 |
| Zurich | 2025 | **75.2** | *41.6** | 73.4 |

## 6. Conclusions

An important observation of our study is the consistency of performance of the TESSERA embedding across the cities regardless of their differing urban morphologies. If reconfirmed over a larger, and even more contrasting set of cities, this would imply that one could also create LCZ maps for cities for which no reference data is available.

The most persistent challenge relates to the 100-m spatial resolution of the WUDAPT LCZ maps used for model training and under-represented LCZ sample classes. While the overall quality of WUDAPT LCZ dataset is sufficiently good to produce informative maps at 10m, higher resolution reference data would not only improve the classification results but also enable a more profound statistical evaluation.

The chosen attention U-Net demonstrated robust performance in mapping LCZ areas with all inputs. The composite S1S2 dataset (combination of Sentinel 1 and Sentinel 2) also performed reasonably well compared to the embeddings in all experiments. As expected, all models struggled to classify the rare classes with very few samples. Higher resolution reference datasets and the combined classification of a larger set of cities would probably be helpful to gather more samples of the rare classes.

The temporal transfer results were encouraging while re-confirming known issues. Models trained on 2024 data produced reasonable LCZ maps for 2025 without any retraining. Even though the 2025 maps did not fully match same-year maps' spatial details, the spatial structure of each city was still well captured. However, both the composites as well as the embeddings reflect year-to-year changes in weather-driven land surface phenology. One should thus either account for those drifts, or train models on multiple-year datasets. With respect to seasonal shifts, S1S2 showed a remarkable stability, whereas the embeddings capture by design much of the year-to-year variation in LSP.

Overall, the results highlight the strong potential of embeddings for global scale LCZ mapping, offering rich spatial and temporal information. TESSERA-based models performed well in different experiment settings, due to their rich spatial and temporal information encoded. Contrary to expectations, the patch-based AlphaEarth embeddings did not surpass the pixel-based TESSERA and Sentinel data in the texturally and structurally rich natural environments.

There is currently no standardized accepted workflow that can apply satellite embedding datasets as the embedding datasets are a new emerging technology. By leveraging the WUDAPT LCZ labels to train the models, the study overcomes the intensive training data preparation and makes LCZ mapping more accessible. To spur innovation and uptake, all the required software and datasets applied in this study are made free and publicly available. Besides enabling reproducibility, this allows user to use the proposed workflow and to test its broader applicability for a larger time (2017 to 2025) as well as contrasting urban environments.

In this paper, we seek to fill the need for the high resolution LCZ mapping framework without intensive ground truth data collection and potentially global scale LCZ registry for sustainable urban planning. We assessed the feasibility and reliability of this deep learning framework by not limiting its application to a single city and single year, as our cross-city analysis included five Swiss major cities under two study periods.

The work is also further scalable as the work is solely based on open-source datasets. Such scalability has allowed us to create a cross-city benchmark. Additionally, the study also provides the performance comparison of one traditional satellite dataset and two major embedding datasets which would enable multiple use cases in the future, e.g. understanding the effectiveness of satellite-derived embeddings.

**Acknowledgments**

Htet Yamin Ko Ko was supported by an ISSI Fellowship. The authors thank Prof. Dr. Michael Rast (International Space Science Institute, Bern) for his valuable comments and support during this research. The authors also thank Dr. Moritz Burger and Prof. Dr. Stefan Brönnimann (Climatology, Geographical Institute, University of Bern) for providing the dataset and for their constructive feedback.

## Appendix: LCZ-Wise Area Computation

Rare LCZ classes (LCZ 3, C, and F) are nearly absent from predictions across all cities even though small areas presented in reference LCZ maps. LCZ 9 (Sparsely built ) exhibits the most severe and consistent underestimation across cities (**Error! Reference source not found.**). LCZ D (Low plants) is constantly overestimated in all cities by all the developed model. LCZ 2 (Compact Midrise) and LCZ G (Water bodies) are performed well by all the developed models (Table 11). Overall, models performed well on Basel, and results of stated cities show overall strong agreement with their reference maps (Table 11). All results (from both Exp I and III) are underestimated in Lausanne.

*Table 11 Per-class area estimates using different models when trained on 2024 WUDAPT data and either evaluated against same year data (2024) or the year after (2025). Unit is in Squared Kilometer. In blue the WUDAPT reference information. Amongst the three models per class, we highlight in* ***bold*** *the most stable 2025 estimation. Please note that study areas with partial data gaps are presented with * and italic font.*

| City | Model | LCZ 1 | LCZ 2 | LCZ 3 | LCZ 4 | LCZ 5 | LCZ 6 | LCZ 7 | LCZ 8 | LCZ 9 | LCZ 10 | LCZ A | LCZ B | LCZ C | LCZ D | LCZ E | LCZ F | LCZ G |
|---|---|---|---|---|---|---|---|---|---|---|---|---|---|---|---|---|---|---|
| Basel | WUDAPT | n.a. | 5.04 | 0.60 | n.a. | 3.17 | 142.75 | n.a. | 17.73 | 6.37 | n.a. | 202.38 | 151.02 | 0.49 | 158.04 | 0.66 | 0.17 | 7.09 |
| | S1S2_2024 | | 5.29 | 0.02 | | 2.27 | 124.33 | | 18.24 | 2.70 | | 212.07 | 112.91 | 0.00 | 209.12 | 0.42 | 0.00 | 8.15 |
| | S1S2_2025 | | 3.73 | **0.00** | | 3.59 | **128.00** | | **16.16** | 1.75 | | 191.39 | **122.48** | **0.00** | **220.03** | **0.42** | 0.00 | 7.95 |
| | AlphaEarth_2024 | | 4.72 | 0.23 | | 1.86 | 127.63 | | 20.83 | 0.18 | | 184.39 | 191.90 | 0.11 | 154.78 | 0.54 | 0.00 | 8.32 |
| | AlphaEarth_2025 | | 5.19 | 0.00 | | 3.98 | 99.88 | | 31.08 | **0.01** | | **192.89** | 173.31 | 0.00 | 180.66 | 0.35 | 0.00 | 8.16 |
| | TESSERA_2024 | | 5.22 | 0.11 | | 1.64 | 136.96 | | 23.67 | 1.52 | | 210.64 | 128.47 | 0.11 | 178.86 | 0.54 | 0.00 | 7.78 |
| | TESSERA_2025 | | **5.44** | 0.01 | | **0.90** | 119.62 | | 30.69 | 0.95 | | 220.29 | 94.45 | 0.02 | 213.81 | 1.70 | 0.00 | **7.62** |
| Bern | WUDAPT | | 1.31 | n.a. | | 5.85 | 36.04 | | 2.14 | 1.16 | | 40.15 | 40.22 | 0.05 | 30.55 | 0.00 | n.a. | 1.50 |
| | S1S2_2024 | | 1.00 | | | 3.21 | 36.33 | | 1.74 | 0.05 | | 40.93 | 29.87 | 0.00 | 44.74 | 0.00 | | 1.15 |
| | S1S2_2025 | | **0.84** | | | **2.19** | **37.93** | | **1.66** | 0.41 | | 35.40 | **33.76** | 0.00 | **45.66** | **0.00** | | **1.15** |
| | AlphaEarth_2024 | | 1.46 | | | 4.20 | 33.57 | | 2.15 | 0.25 | | 40.04 | 38.64 | 0.00 | 36.97 | 0.00 | | 1.75 |
| | AlphaEarth_2025 | | 1.91 | | | 6.28 | 21.60 | | 5.13 | 0.01 | | 33.49 | 45.88 | 0.01 | 42.44 | 0.02 | | 2.25 |
| | TESSERA_2024 | | 1.19 | | | 4.56 | 36.34 | | 2.15 | 0.02 | | 42.39 | 35.04 | 0.00 | 35.92 | 0.00 | | 1.22 |
| | TESSERA_2025 | | 0.50 | | | 0.02 | 40.53 | | 2.25 | **0.11** | | **46.85** | 17.77 | 0.00 | 49.60 | 0.01 | | 1.39 |
| Geneva | WUDAPT | | 7.18 | 0.05 | | 5.11 | 164.51 | | 11.18 | 26.02 | | 205.41 | 157.32 | 0.07 | 236.21 | 1.65 | 0.29 | 87.06 |
| | S1S2_2024 | | 6.48 | 0.00 | | 2.53 | 131.27 | | 11.67 | 22.26 | | 200.26 | 118.95 | 0.00 | 319.29 | 0.77 | 0.00 | 88.58 |
| | S1S2_2025 | | 6.01 | 0.00 | | **1.84** | **123.84** | | **12.93** | 19.39 | | 183.14 | **119.81** | 0.00 | **346.08** | 1.18 | 0.00 | 87.84 |
| | AlphaEarth_2024* | | *7.22* | *0.00* | | *3.31* | *137.98* | | *12.11* | *4.51* | | *118.72* | *125.08* | *0.03* | *215.60* | *1.14* | *0.01* | *87.77* |
| | AlphaEarth_2025* | | *7.22* | *0.00* | | *4.90* | *130.94* | | *16.27* | *2.93* | | *115.02* | *110.06* | *0.02* | *235.59* | *1.55* | *0.01* | *89.13* |
| | TESSERA_2024 | | 7.69 | 0.00 | | 2.27 | 170.55 | | 12.62 | 21.71 | | 214.21 | 141.40 | 0.00 | 240.86 | 2.00 | 0.00 | 88.74 |
| | TESSERA_2025 | | **7.24** | 0.00 | | 3.43 | 147.18 | | 14.09 | 2.32 | | **228.85** | 116.05 | 0.00 | 291.70 | **1.75** | 0.00 | **89.44** |
| Lausanne | WUDAPT | | 1.85 | n.a. | | 2.93 | 70.40 | | 7.68 | 15.78 | | 96.81 | 91.81 | 0.15 | 280.81 | 0.20 | 0.13 | 167.12 |
| | S1S2_2024 | | 1.81 | | | 1.88 | 64.68 | | 6.07 | 5.13 | | 103.07 | 63.97 | 0.00 | 321.07 | 0.00 | 0.01 | 167.95 |
| | S1S2_2025 | | 1.41 | | | **1.53** | 60.99 | | **6.23** | **5.73** | | 92.19 | 75.54 | **0.00** | 324.22 | **0.00** | 0.00 | **167.82** |
| | AlphaEarth_2024 | | 1.70 | | | 2.14 | 58.95 | | 7.30 | 9.18 | | 94.71 | 93.22 | 0.11 | 301.09 | 0.09 | 0.00 | 167.16 |
| | AlphaEarth_2025 | | 3.30 | | | 3.09 | 56.51 | | 10.80 | 3.51 | | **97.71** | **89.55** | 0.16 | **303.18** | 0.10 | 0.00 | 167.73 |
| | TESSERA_2024 | | 1.62 | | | 2.37 | 69.49 | | 8.27 | 6.24 | | 103.04 | 75.66 | 0.01 | 301.48 | 0.10 | 0.00 | 167.36 |
| | TESSERA_2025 | | **1.75** | | | 1.40 | **68.39** | | 9.14 | 2.98 | | 112.14 | 55.94 | 0.00 | 316.03 | 0.15 | 0.00 | 167.72 |
| Zurich | WUDAPT | | 6.33 | 0.16 | | 11.13 | 168.35 | | 16.23 | 4.34 | | 172.52 | 161.73 | 0.05 | 137.47 | 1.7 | 0.05 | 28.67 |
| | S1S2_2024 | | 8.48 | 0.02 | | 5.41 | 157.08 | | 12.08 | 0.41 | | 177.22 | 107.18 | 0.00 | 210.90 | 1.54 | 0.00 | 28.41 |
| | S1S2_2025 | | **7.68** | **0.04** | | **5.32** | **157.07** | | **12.34** | 0.62 | | **171.76** | **113.70** | 0.00 | **210.01** | **1.72** | 0.00 | **28.48** |
| | AlphaEarth_2024* | | *6.67* | *0.03* | | *11.57* | *81.17* | | *23.40* | *0.00* | | *105.15* | *318.60* | *0.00* | *98.63* | *1.03* | *0.00* | *9.90* |
| | AlphaEarth_2025* | | *11.93* | *0.00* | | *42.41* | *97.92* | | *27.19* | *0.00* | | *91.11* | *275.93* | *0.00* | *92.00* | *1.34* | *0.00* | *16.12* |
| | TESSERA_2024 | | 6.09 | 0.03 | | 9.84 | 160.31 | | 16.88 | 0.06 | | 171.21 | 172.82 | 0.00 | 140.50 | 1.92 | 0.00 | 29.08 |
| | TESSERA_2025 | | 10.27 | 0.00 | | 4.80 | 152.72 | | 18.86 | **0.12** | | 193.48 | 102.75 | 0.00 | 193.76 | 2.51 | 0.00 | 28.93 |